\documentclass[sigconf]{acmart}

\AtBeginDocument{%
  }

\setcopyright{acmlicensed}
\copyrightyear{2026}
\acmYear{2026}
\setcopyright{cc}
\setcctype{by}
\acmConference[MM '26] {Proceedings of the 34th ACM International Conference on Multimedia}{November 10--14, 2026}{Rio de Janeiro, Brazil.}
\acmBooktitle{Proceedings of the 34th ACM International Conference on Multimedia (MM '26), November 10--14, 2026, Rio de Janeiro, Brazil}
\acmISBN{979-8-4007-2213-4/2026/11}
\acmDOI{10.1145/3767308.3836415}

\usepackage{subcaption}
\usepackage{multicol}
\usepackage{multirow}
\usepackage{tabularx}
\usepackage{booktabs}

\begin{document}

\title{Rule-Compliant Visual Spatial Planning for Multimodal Large Language Models}

\author{Yu Chen}
\orcid{0009-0008-3382-3405}
\affiliation{%
  \institution{Wangxuan Institute of Computer Technology, Peking University}
  \city{Beijing}
  \country{China}}
\email{yu_chen@stu.pku.edu.cn}

\author{Ting Lei}
\orcid{0000-0001-9452-577X}
\affiliation{%
  \institution{Wangxuan Institute of Computer Technology, Peking University}
  \city{Beijing}
  \country{China}}
\email{ting_lei@pku.edu.cn}

\author{Yaoyi Li}
\orcid{0000-0002-6579-8274}
\affiliation{%
  \institution{Yinwang Intelligent Technology Co., Ltd}
  \city{Shenzhen}
  \country{China}}
  \email{liyaoyi@yinwang.com}

\author{Jia Cai}
\orcid{0000-0002-8796-961X}
\affiliation{%
  \institution{Yinwang Intelligent Technology Co., Ltd}
  \city{Shenzhen}
  \country{China}}
  \email{caijianwpu@gmail.com} 

\author{Zhecen Wu}
\orcid{0009-0007-0071-554X}
\affiliation{%
  \institution{Yinwang Intelligent Technology Co., Ltd}
  \city{Shenzhen}
  \country{China}}
  \email{wuzhecen@yinwang.com}

\author{Yang Liu}
\orcid{0000-0002-4259-3882}
\authornote{Corresponding author.}
\affiliation{%
  \institution{Wangxuan Institute of Computer Technology, Peking University}
  \city{Beijing}
  \country{China}}
\email{yangliu@pku.edu.cn}

\renewcommand{\shortauthors}{Yu Chen et al.}

\begin{abstract}

Multimodal large language models (MLLMs) combine linguistic reasoning with visual perception, yet their ability to perform visual spatial planning under explicit or previously unseen rule constraints remains underexplored. This setting requires models to jointly understand spatial layouts, interpret natural-language rules, and plan valid actions accordingly.
To address this gap, we introduce RuleMaze, a controllable benchmark in which MLLMs must navigate mazes while obeying natural-language rules of varying complexity. RuleMaze isolates rule-compliant spatial planning by requiring accurate perception, rule interpretation, and constrained action planning. To enable scalable and systematic rule construction, we propose Language–Logic–Function Hybridization, which automatically generates natural-language rules and translates them into logical representations and executable validators, eliminating manual rule engineering.
To improve rule following and generalization, we introduce Disentangled Multimodal Planning (DMP), which separates perception, execution, and rule verification through interpretable reasoning primitives. By disentangling these components, DMP facilitates systematic generalization to more complex and previously unseen rules, while providing transparent intermediate planning traces.
Experiments demonstrate that DMP substantially improves rule compliance and planning success compared to end-to-end textual planning baselines. Overall, RuleMaze establishes a principled benchmark for studying grounded and interpretable rule-based spatial planning in MLLMs. Code is available at https://github.com/oceanflowlab/RuleMaze.

\end{abstract}





\maketitle

\section{Introduction}

Recently, multimodal large language models (MLLMs) have rapidly advanced beyond traditional language-only models by incorporating native visual inputs~\cite{achiam2023gpt4report,peng2026survey,team2023gemini,mo2026distilling,bai2025qwen2-5vl}. Through large-scale multimodal pretraining, these models have achieved impressive performance on tasks such as image captioning~\cite{yang2023exploring,hu2023promptcap,liu2026confidence}, visual question answering~\cite{shao2023prompting,yin2025toolvqa,lei2026unleashing}, and general visual reasoning~\cite{xu2025trkt,ying2024mmt,gao2026taming}. 
Recent works have begun to explore spatial planning and multi-step action reasoning in visual environments, which constitutes a necessary step toward embodied intelligence~\cite{tang2025lego,MVoT,ting2025hoi,xu2025visualplanningletsthink}. 


However, many of these settings assume unconstrained planning or optimize for a single objective without requiring explicit compliance with externally specified rules.
A particularly important yet understudied capability is rule-compliant spatial planning: the ability to reason over visual spatial layouts, interpret explicit rules expressed in natural language, and generate multi-step action plans that satisfy those rules. This capability is central to many real-world settings, such as autonomous driving~\cite{tian2024drivevlm,ma2024dolphins,yang2026gala} and embodied intelligence~\cite{yang2025ar,chang2024lgmcts,chen2025helmet,yang2025planllm}, where agents must operate under changing, context-specific constraints rather than fixed global objectives.
For example, an autonomous agent deployed in a new country may encounter traffic regulations that differ from those seen during training, requiring it to correctly interpret and obey previously unseen rules rather than relying on memorized behaviors. Despite its importance, current benchmarks and analyses offer little systematic evaluation of how well MLLMs can perform such constrained spatial planning.

Studying rule-compliant spatial planning in realistic environments presents substantial challenges. First, from a data perspective, real-world scenes rarely provide ground-truth action sequences corresponding to dynamically varying rule sets. Collecting labeled demonstrations for each new combination of environment and rule is prohibitively expensive, making it difficult to scale evaluation or training across diverse rule constraints.
Second, from a methodological perspective, rule-compliant spatial planning requires the tight integration of multiple competencies: accurate visual perception, long-horizon spatial planning, precise rule interpretation, and strict adherence to rules during execution. End-to-end formulations that entangle these capabilities within a single model often struggle to coordinate them reliably, leading to brittle behavior when confronted with multi-step constraints or previously unseen rules.
These limitations motivate the need for (i) a controllable evaluation environment that supports systematic variation of rule complexity, and (ii) reasoning frameworks that explicitly structure how MLLMs perceive, execute, and verify actions under constraints. Addressing these needs is essential for understanding the limits of current MLLMs and for developing models that can generalize to complex, rule-governed spatial planning tasks.

\begin{figure}[ht]
    \centering 
    \begin{subfigure}[b]{0.49\textwidth}
        \centering
        \includegraphics[width=\textwidth]{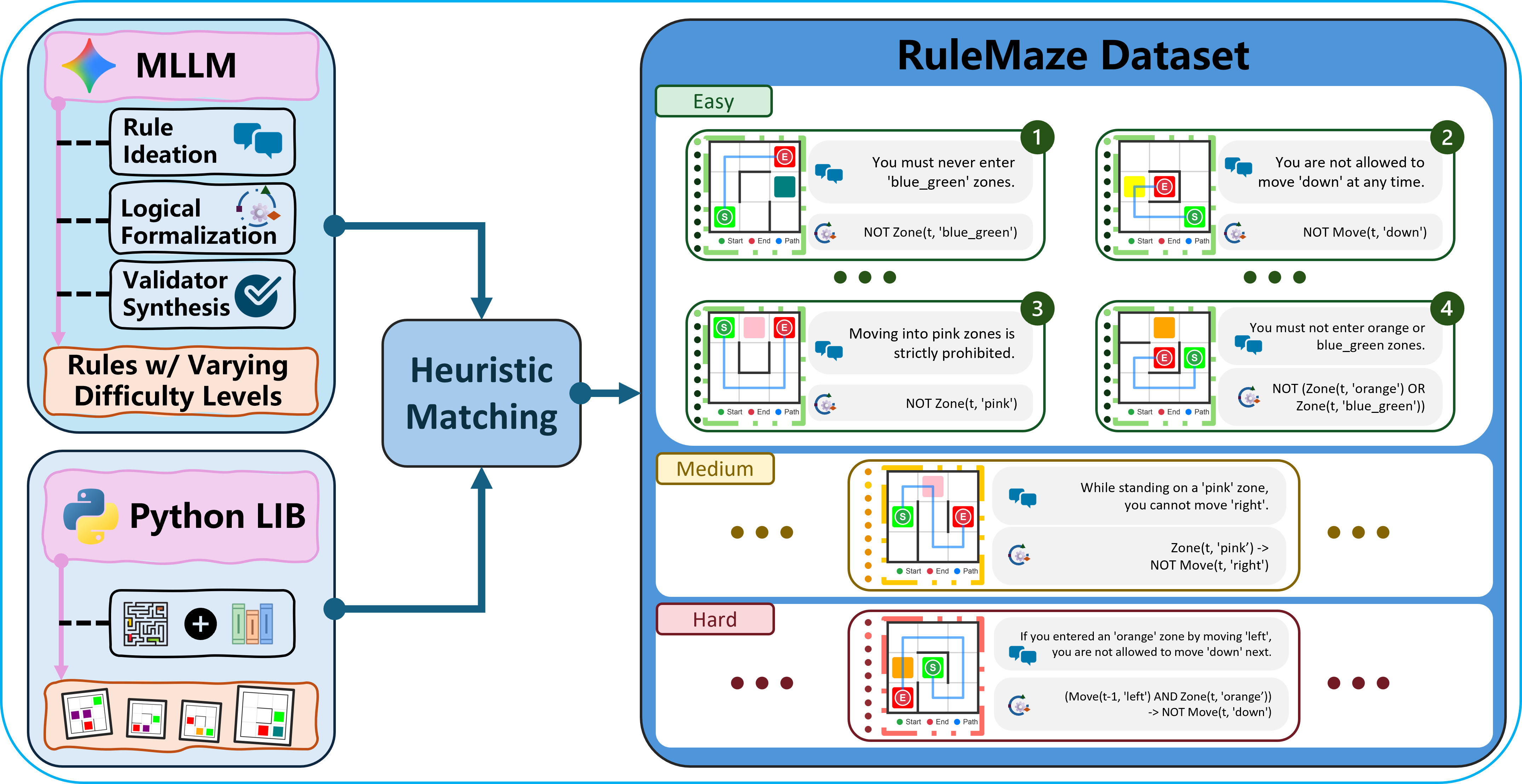}
        \caption{RuleMaze dataset construction.}
        \label{fig:teaser_a}
    \end{subfigure}
    \hfill
    \begin{subfigure}[b]{0.49\textwidth}
        \centering
        \includegraphics[width=\textwidth]{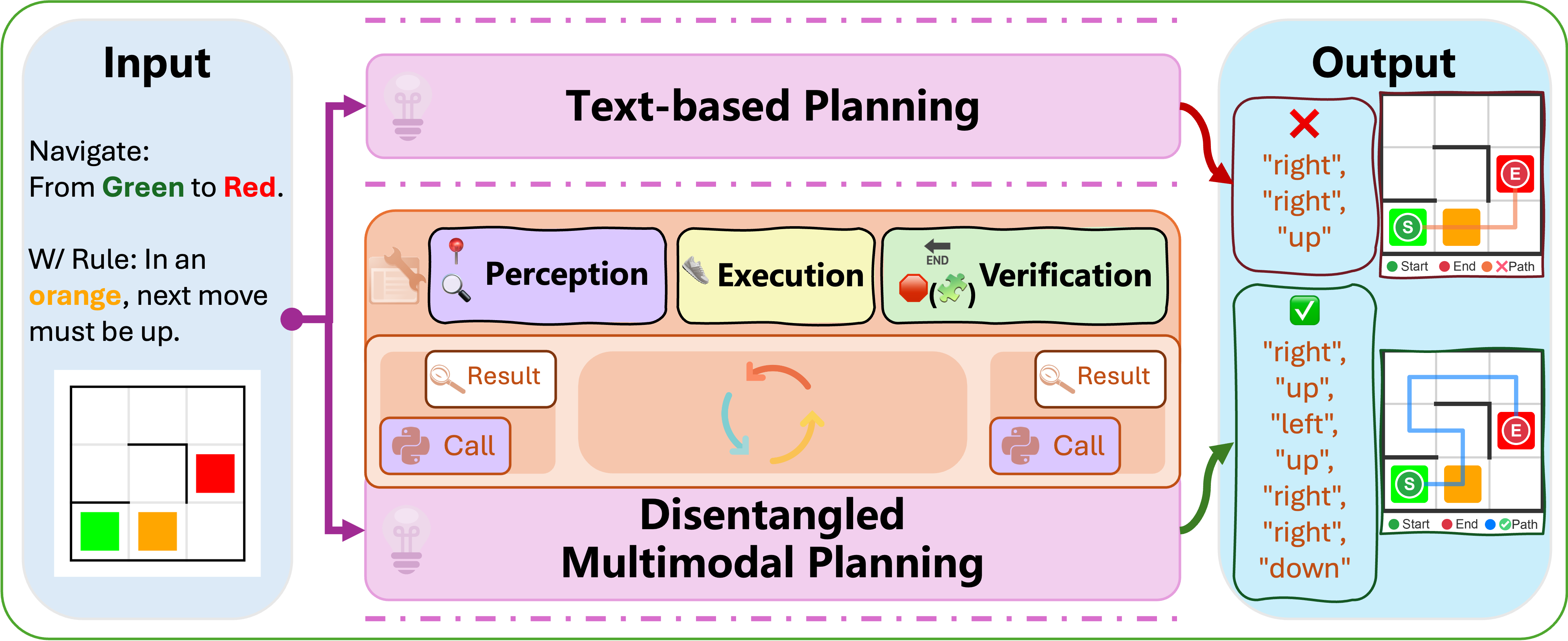} 
        \caption{Disentangled Multimodal Planning (DMP) framework.}
        \label{fig:teaser_b}
    \end{subfigure}
    \caption{Overview of the RuleMaze dataset and the proposed Disentangled Multimodal Planning (DMP) framework.}
    \label{fig:teaser}
\end{figure}

To address this gap, we introduce RuleMaze, a controllable benchmark for evaluating visual spatial planning under explicit natural-language rules. As shown in Fig.~\ref{fig:teaser_a}, RuleMaze pairs diverse maze-based visual environments with rule specifications that constrain which multi-step action plans are considered valid.
To construct scalable rule spaces with controlled complexity, we propose a Language--Logic--Function Hybridization pipeline that transforms free-form rule concepts into executable constraints. Specifically, LLMs are used for \emph{Rule Ideation} to generate natural-language rules, which are then translated via \emph{Logical Formalization} into structured logical representations. We further perform \emph{Validator Synthesis} to automatically produce executable functions that verify whether a candidate action trajectory satisfies a given rule, enabling precise and automated evaluation without manual rule engineering.
In parallel, we procedurally generate a large collection of maze images with diverse layouts and visual configurations using a Python-based library.
Each maze is designed to admit multiple feasible routes from the start to the goal.
To ensure rules are evaluated only in applicable environments, we introduce a heuristic rule--maze matching strategy: for each rule, we select mazes in which the rule admits a non-empty set of valid trajectories while inducing a unique rule-compliant solution among multiple distractor routes.
To support controlled analysis of reasoning difficulty, rules are categorized into easy, medium, and hard levels based on the syntactic complexity of their logical formalizations, measured by the number of logical connectives. Overall, RuleMaze offers (i) diverse procedurally generated mazes with multiple competing routes, (ii) systematically constructed rule sets spanning multiple difficulty levels, and (iii) unambiguous action-level trajectories under explicit constraints, making it a principled benchmark for studying rule-based spatial planning in MLLMs.

Beyond benchmarking, we propose Disentangled Multimodal Planning (DMP), a general framework for improving rule-compliant spatial planning in MLLMs, as illustrated in Fig.~\ref{fig:teaser_b}. Unlike end-to-end training, which entangles perception, execution, and rule reasoning into a single data-driven process and requires substantial supervision to relearn these capabilities whenever rules change, DMP explicitly modularizes rule-constrained problem solving and trains the model to coordinate reusable capabilities.
Specifically, DMP decomposes the task into three interpretable components—visual perception, action execution, and rule verification—each implemented as a set of callable tools. Given a maze image and a natural-language rule, the model alternates between internal reasoning and external tool invocation, deciding when to perceive visual evidence, propose candidate actions, and verify rule compliance. Importantly, having access to all tools does not guarantee solving the problem: the model still needs to learn in a data-driven way when and how to invoke each tool to successfully complete rule-constrained planning. DMP therefore focuses training on the core controller’s ability to select, sequence, and integrate tool outputs.
Crucially, rule verification is externalized through dedicated validator tools automatically synthesized from logical rule representations. These validators are fully replaceable, allowing new or previously unseen rules to be incorporated without retraining the controller. 

In summary, this work makes the following contributions.
(1) We introduce RuleMaze, a controllable benchmark for evaluating spatial planning under explicit natural-language rules, featuring diverse maze environments and rule–maze applicability grounded by executable verification tools.
(2) We propose a Language–Logic–Function Hybridization pipeline that enables scalable construction of rule sets with controlled complexity, transforming natural-language rules into logical representations and executable validators without manual rule engineering.
(3) We present Disentangled Multimodal Planning (DMP), a framework that enhances rule-following behavior in MLLMs by equipping them with explicit perception, execution, and verification tools, enabling multi-step, interpretable reasoning under constraints.
(4) Through extensive experiments, we demonstrate that DMP substantially improves rule compliance and planning performance compared to end-to-end textual planning, especially under more complex and previously unseen rules.

\section{Related work}

\subsection{Spatial Planning}

Planning has long been central in AI, traditionally relying on formal representations and algorithms~\cite{sutton1991planning,guo2014deep,aeronautiques1998pddl}, typically in predefined and constrained settings. With recent advances in large language models (LLMs), it is now compelling to explore whether LLMs—viewed as general-purpose intelligent agents—can perform planning across diverse environments~\cite{stechly2024self,kambhampati2024llms,kambhampati2024can,shao2026chinatravel}. For instance, AlphaMaze~\cite{AlphaMaze} converts maze navigation into a tokenized textual task, bridging language models with spatial planning in a purely symbolic setting. However, such approaches are limited to abstract, text-only decision-making, without observing visual state changes.
Building on these capabilities, multimodal language models (MLLMs) extend planning into visually grounded spatial reasoning, where actions dynamically alter the environment. Benchmarks and methods include VisualCoT~\cite{du2025revisitingvisualCoT}, which studies chain-of-thought designs for generalizable spatial planning; Ariadne~\cite{shen2025ariadne}, which applies reinforcement learning with verified rewards and a difficulty-aware curriculum; MVoT~\cite{MVoT}, which integrates textual reasoning with visually grounded mental imagery to enhance interpretability and robustness; and VPRL~\cite{xu2025visualplanningletsthink}, which conducts planning entirely through sequences of images, enabling models to "think" directly in the visual domain.

Unlike prior work that focuses on abstract planning or visual reasoning alone, our approach emphasizes rule-compliant spatial planning: natural-language rules explicitly constrain the action space, enabling a controllable and verifiable framework for evaluating and improving rule-based spatial planning in MLLMs.

\subsection{Rule and Knowledge Injection}

Rule and knowledge injection techniques enhance model reasoning by incorporating external constraints or factual information, either dynamically at inference time or statically through parameter adaptation.
Dynamic injection incorporates constraints at inference time to guide model behavior. For example, in task-oriented reasoning, RAP~\cite{hao2025rap} retrieves personalized knowledge rules to shape response generation. PokeMQA~\cite{gu2024pokemqa} handles knowledge editing in multi-hop QA by decoupling reasoning chains from conflict detection, ensuring updated knowledge is propagated consistently. DeepEdit~\cite{wang2024deepedit} frames rule-based knowledge editing as constrained decoding, iteratively validating reasoning steps against injected rules.
Static injection embeds constraints directly into model parameters. VisEdit~\cite{chen2025visedit} edits intermediate visual representations to enforce updated visual knowledge affecting downstream predictions. 
RECT~\cite{gu2024RECT} addresses the side effects of knowledge editing on general capabilities and proposes regularization to constrain weight updates, preventing overfitting to edited knowledge.

While existing methods have demonstrated the effectiveness of rule and knowledge injection for textual reasoning, they primarily focus on influencing the language output space. In contrast, spatial planning requires rules that directly constrain the decisions in the action space. Moreover, existing methods do not address sequential planning or visually grounded decision-making, whereas our approach enables models to generate action sequences that are both executable and compliant with explicit rules.

\section{RuleMaze Dataset Construction}

\begin{figure*}[htbp]
    \centering
    \includegraphics[width=0.85\textwidth]{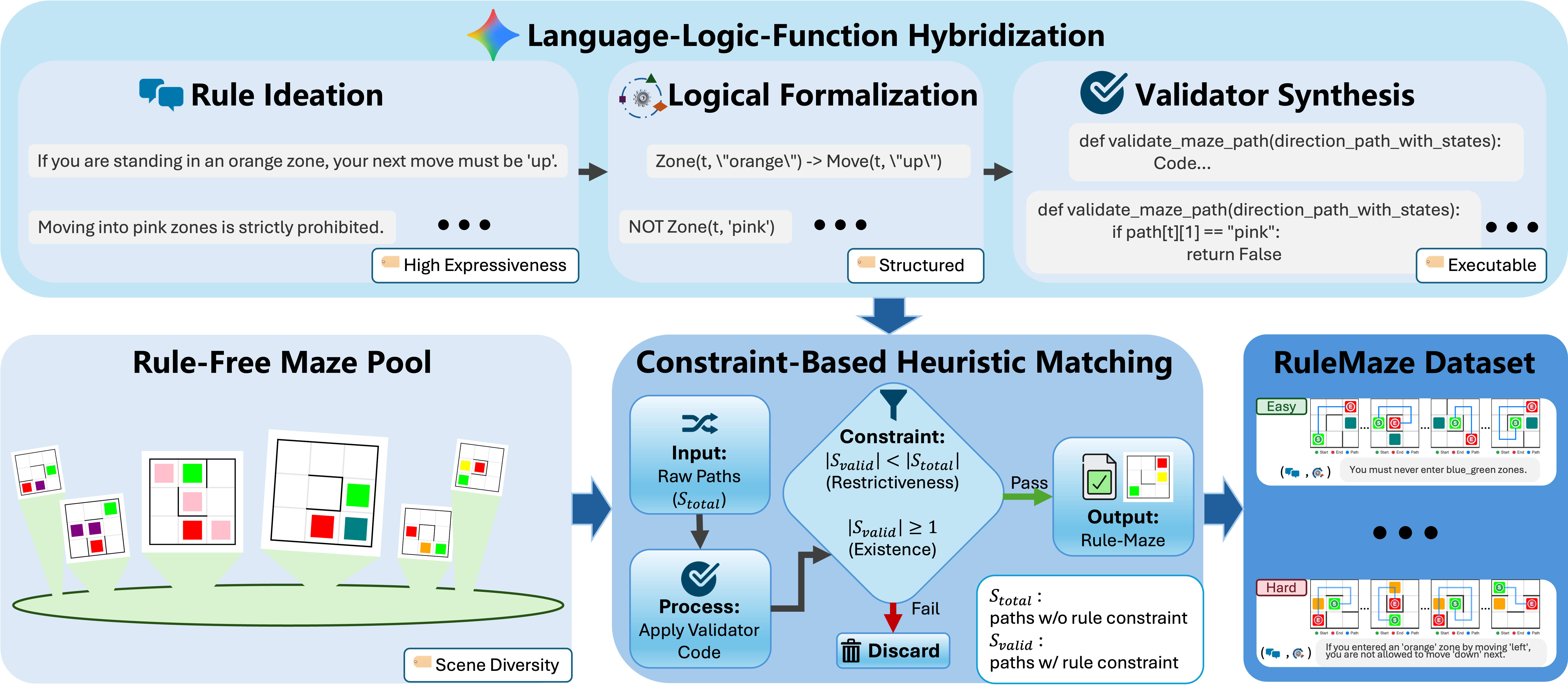}
    \caption{Overview of the RuleMaze construction pipeline. Free-form rule concepts are generated via Rule Ideation, translated into structured representations through Logical Formalization, and compiled into executable validators by Validator Synthesis. In parallel, diverse maze environments are procedurally generated. A constrained-based heuristic rule–maze matching strategy selects applicable mazes for each rule, yielding rule–maze pairs with unambiguous rule-compliant trajectories.}
    \label{fig:Rulemaze}
\end{figure*}

Investigating rule-compliant spatial planning in realistic settings is fundamentally challenged by the scarcity of annotated data, as real-world environments rarely offer ground-truth trajectories corresponding to dynamic rule sets.
To address this, we present RuleMaze, a procedurally generated dataset of rule–maze pairs with executable validators. Our pipeline uses a Language–Logic–Function framework to generate rules with controlled complexity. In parallel, we procedurally synthesize diverse maze layouts. Through rule–maze matching, we pair each rule exclusively with mazes where it yields a non-empty set of compliant trajectories with a unique rule-consistent solution, while challenging distractor trajectories still exist. RuleMaze thus enables controlled and scalable evaluation for rule-constrained spatial planning research.

\subsection{Language--Logic--Function Hybridization}
\label{subsec:Hybridization}

The Language--Logic--Function Hybridization pipeline bridges expressive natural-language rules and precise, machine-checkable constraints, as illustrated in the upper part of Fig.~\ref{fig:Rulemaze}. It comprises three stages: Rule Ideation, Logical Formalization, and Validator Synthesis.

\paragraph{Rule Ideation.}
In the first stage, we use large language models (LLMs) to generate diverse and linguistically rich natural-language rules that constrain planning behavior. These rules go beyond simple goal specifications and capture conditional and state-dependent constraints---for example, \textit{``Moving into pink zones is strictly prohibited''} or \textit{``If you are standing in an orange zone, your next move must be up.''} This stage prioritizes compositional expressiveness, producing rules that demand non-trivial interpretation and multi-step reasoning.

\paragraph{Logical Formalization.}
Each natural-language rule is then translated into a structured logical representation that formally encodes its semantics. For instance, the rule \textit{``If you are standing in an orange zone, your next move must be up''} is formalized as
\(
\mathtt{Zone}(t, \mathrm{orange}) \rightarrow \mathtt{Move}(t, \mathrm{up}),
\)
while \textit{``Moving into pink zones is strictly prohibited''} is expressed as
\(
\lnot \mathtt{Zone}(t, \mathrm{pink}).
\)
These logical forms provide a compact and structured intermediate representation that supports both semantic analysis and automated execution.

\paragraph{Validator Synthesis.}
Finally, we automatically synthesize an executable validator from each logical representation using LLMs. Specifically, the LLM generates a Python function that takes a candidate trajectory as input and determines whether it complies with the corresponding rule. By decomposing complex rules into simpler logical components, validator synthesis becomes more reliable and scalable. Empirically, the generated validators achieve 93.3\% correctness before manual verification. To further ensure dataset quality, all validators are manually inspected and verified before deployment.

\subsection{Constraint-Based Heuristic Matching}
\label{subsec:matching}

To support scalable dataset construction, we first generate a large, rule-agnostic maze pool using a procedural generator. These mazes exhibit diversity in spatial layout, forming a rich environmental basis without prior rule constraints.

Given this pool, we introduce a constraint-based heuristic matching strategy to associate each rule with mazes where it is meaningfully evaluable, as illustrated in the lower-middle part of Fig.~\ref{fig:Rulemaze}. This process ensures that each rule--maze pair admits solvable yet non-trivial planning tasks.

For a given rule, candidate mazes are those containing the symbolic elements referenced by the rule (e.g., specific colored zones). Within each candidate maze, we enumerate all feasible start-to-goal trajectories, denoted $S_\text{total}$, without considering rule constraints. Applying the rule's validator (from Section~\ref{subsec:Hybridization}) yields the subset of compliant trajectories $S_\text{valid} \subseteq S_\text{total}$.
A maze is retained for a rule if it satisfies the following two constraints: (1) Restrictiveness: $|S_\text{valid}| < |S_\text{total}|$, ensuring the rule meaningfully restricts behavior. (2) Existence: $|S_\text{valid}| \geq 1$, 
guaranteeing at least one rule-compliant solution exists.

Rule–maze pairs satisfying both constraints are accepted; otherwise, the maze is discarded for that rule. 
This matching strategy produces rule--maze pairs with clear behavioral constraints, enabling precise evaluation of rule-compliant spatial planning in multimodal agents.

\begin{figure}[htbp]
    \centering
    \includegraphics[width=0.5\textwidth]{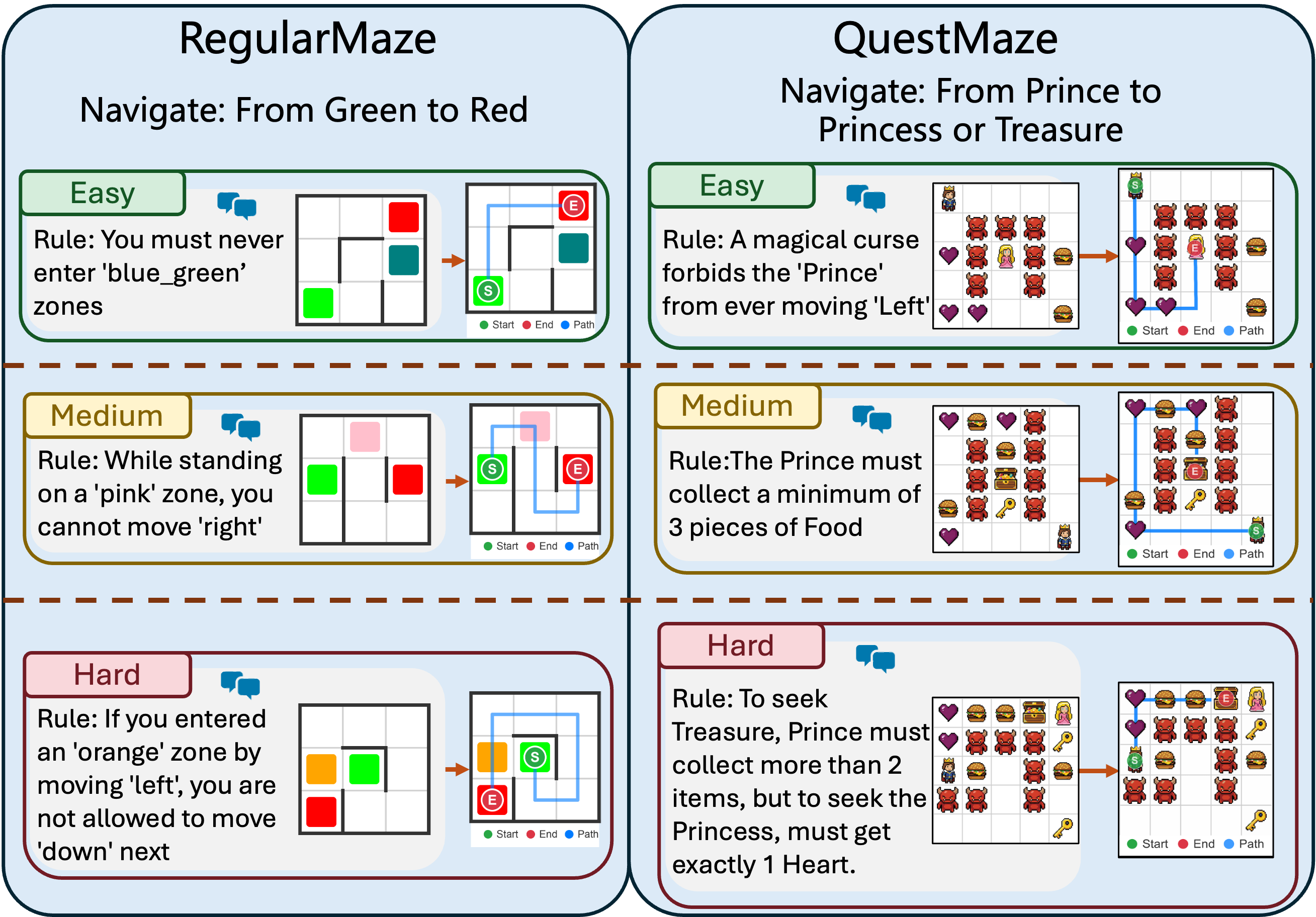}
    \caption{RuleMaze dataset illustration under both the RegularMaze and QuestMaze scenarios.}
    \label{fig:dataset-samples}
    \vspace{-2em}
\end{figure}

\subsection{RuleMaze Dataset}
\label{subsec:RuleMaze}

Through the Language–Logic–Function Hybridization and the Constraint-based Heuristic Matching procedure described above, we construct RuleMaze, a large-scale dataset of rule–maze pairs for evaluating rule-compliant spatial planning in MLLMs. The dataset comprises two complementary scenarios: RegularMaze and QuestMaze.

\paragraph{RegularMaze.}
In the RegularMaze scenario, the start and goal locations are indicated by a green and a red cell, respectively. Rules reference an additional colored zone (e.g., orange or pink), while all remaining cells are visually uniform white. This setting emphasizes spatial reasoning under static goals and local constraints, requiring models to interpret color-based rules while planning globally optimal paths.

\paragraph{QuestMaze.}
QuestMaze introduces a more semantically rich and challenging environment. The start location is represented by a prince icon, while the goal is a princess or treasure icon. Rules in this scenario involve symbolic item icons (e.g., keys) that may modify planning constraints or affect the goal condition. All other cells are rendered with a grass-textured background. 
Compared to RegularMaze, QuestMaze requires reasoning over historical state information—such as whether and how many times an item has been collected—as well as handling dynamically changing target conditions specified by different rules. Together, these factors induce state- and scene-dependent action constraints and substantially increase planning complexity.


Example mazes from both scenarios are shown in Fig.~\ref{fig:dataset-samples}.
For each scenario, rules are categorized into easy, medium, and hard levels based on the syntactic complexity of their logical formalizations, measured by the number of logical connectives. 
Overall, Regular Maze isolates core rule-conditioned spatial planning, while QuestMaze further challenges models with memory-dependent rules and non-stationary goals. Together, these scenarios enable fine-grained evaluation of MLLMs’ ability to perform interpretable, multi-step, and rule-compliant spatial planning across varying levels of difficulty.

\section{Disentangled Multimodal Planning}


Rule-compliant visual spatial planning requires the coordination of multiple heterogeneous capabilities, including visual perception, action execution, rule interpretation, and long-horizon reasoning. End-to-end formulations entangle these components into a single data-driven process, forcing the model to relearn largely stable capabilities whenever rules change, resulting in data inefficiency, brittle behavior, and limited diagnostic insight when failures occur.
In contrast, many core functionalities—such as perceiving maze layouts, executing primitive actions, and verifying rule compliance—are reusable across tasks and rules. The central challenge instead lies in learning \emph{when} and \emph{how} to invoke these capabilities under evolving constraints. Motivated by this observation, we introduce Disentangled Multimodal Planning (DMP), a framework that modularizes rule-constrained planning into interpretable and replaceable tools, and trains a controller to reason over intermediate states and selectively orchestrate tool usage. By externalizing stable components and focusing learning on coordination, DMP enables data-efficient training and systematic generalization to novel and previously unseen rules.

\subsection{Task Formulation}

We study rule-constrained visual planning in grid-based maze environments. Each task instance consists of three components: a maze image $\mathcal{I}$, a goal instruction $\mathcal{G}$, and a natural-language rule specification $\mathcal{R}$. The maze image $\mathcal{I}$ depicts a 2D grid environment with a designated start location, a target location, and rule-relevant cells or semantic objects. The goal instruction $\mathcal{G}$ specifies the goal of the task, e.g., “Navigate from Green to Red”. The rule specification $\mathcal{R}$ defines additional constraints on valid actions, such as “In an orange cell, the next move must be up”.

The model is required to produce an action sequence\\
$\mathbf{a} = (a_1, a_2, \ldots, a_T)$, $\quad a_t \in \mathcal{A},$
where $\mathcal{A} = \{\texttt{up}, \texttt{down}, \texttt{left}, \texttt{right}\}$ denotes the action space and $T$ denotes the number of executed movement actions. Executing $\mathbf{a}$ from the start position must (i) reach the target location specified in $\mathcal{I}$ and (ii) satisfy the rule constraints defined by $\mathcal{R}$ at every step.
\renewcommand{\arraystretch}{0.8}
\begin{table*}[htbp]
\centering
\caption{Summary of disentangled tools in the DMP framework.}
\label{tab:tools_summary}
\small
\begin{tabular}{lllll}
\toprule
Category & Tool Name & Inputs & Function Description & Output \\ \midrule
Perception & \texttt{LocateStart} & Image $\mathcal{I}$ & Identifies and marks the initial agent position & Image $I_0$ \\
  & \texttt{InspectGrid} & Image $I_t$ & Identifies semantic symbols at specific cell & Symbol $s_t$ \\ 
\midrule
Execution & \texttt{ExecuteMove} & Image $I_t$, Action $a_t$ & Updates agent position and visual state & Image $I_{t+1}$ \\ 
\midrule
Verification & \texttt{VerifyRule} & $a_{0:t}$, $s_{0:t}$, Rule $R$ & Checks trajectory compliance with rule $R$ & Boolean \\
 & \texttt{VerifyEndpoint} & Image $I_t$ & Checks if agent has reached the target & Boolean \\ \bottomrule
\end{tabular}
\end{table*}

\begin{figure*}[htbp]
    \centering
    \includegraphics[width=0.85\textwidth]{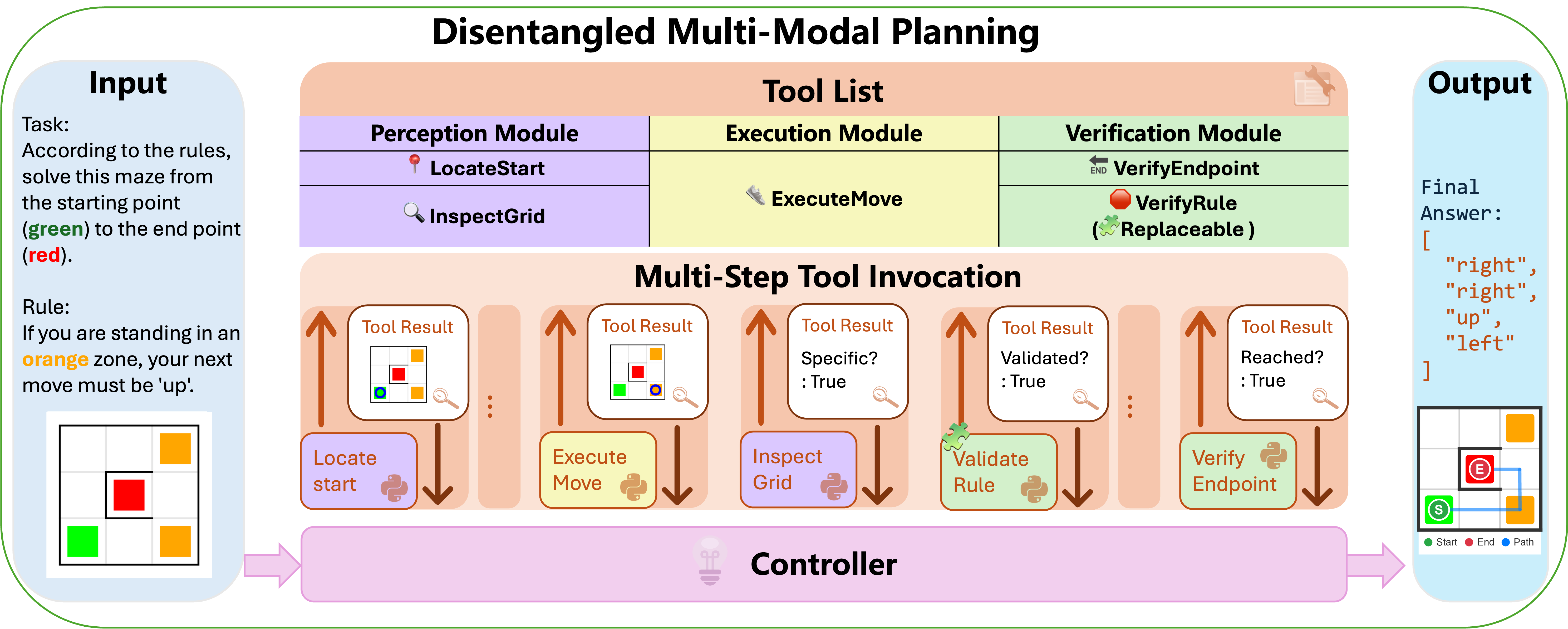}
    \caption{Overall framework of our proposed Disentangled Multi-Modal Planning.}
    \label{fig:method}
    \vspace{-1em}
\end{figure*}

\subsection{Disentangled Tool-Based Planning}

We equip MLLMs with a set of disentangled and interpretable tools, organized into three categories: Perception, Execution, and Verification, as illustrated in Tab.~\ref{tab:tools_summary}. Formally, we define a tool set
$\mathcal{T} = \mathcal{T}_{\text{perc}} \cup \mathcal{T}_{\text{exec}} \cup \mathcal{T}_{\text{ver}},$
where each tool $\tau \in \mathcal{T}$ is a callable function with structured inputs and outputs.
Perception tools ground the model’s reasoning in the maze image, execution tools deterministically update the visual state by applying movement actions, and verification tools assess rule compliance over the resulting trajectories, enabling step-wise, interpretable planning under explicit constraints.

\paragraph{Perception Tools.}

Perception tools extract symbolic information from the maze image. Given an image $I_t$ at step $t$, these tools produce structured observations: $o_t = \tau_{\text{perc}}(I_t),$ where $o_t$ may include the start location, and whether the current cell corresponds to a special symbol (e.g., a colored zone or item).

\paragraph{Execution Tools.}

Execution tools implement environment transitions. Given an action $a_t \in \mathcal{A}$ and the current image $I_t$, the execution tool computes $I_{t+1} = \tau_{\text{exec}}(I_t, a_t),$ which updates the agent's position and returns a new maze image with the current location explicitly marked. This visual state update allows the model to reason over spatial transitions using explicit visual feedback.

\paragraph{Verification Tools.}

Verification tools assess task completion and rule compliance. Given a partial trajectory $a_{0:t}$ and the corresponding states $s_{0:t}$, verification tools return binary signals:\\ $b_{t}^{rule} =
\tau_{\text{ver}}^{rule}(a_{0:t}, s_{0:t}, R)$ and
$b_{t}^{goal} = \tau_{\text{ver}}^{goal}(I_t)$, indicating whether the trajectory satisfies the rule constraints or whether the target has been reached. Each rule $R$ is associated with a dedicated validator $\tau_R \in \mathcal{T}_{\text{ver}}^{rule}$, synthesized offline from its logical formalization by an external LLM~\cite{team2023gemini} before controller training and inference. These validators are replaceable and compositional, enabling the framework to support new rules without retraining the core controller, as illustrated in Fig.~\ref{fig:method}. 

\subsection{Multi-Step Tool Invocation}

Disentangled Multi-Modal Planning proceeds as an iterative decision process over reasoning steps and tool invocations. At each step $t$, the model maintains an internal reasoning state $r_t$, initialized as $r_0$. The input at step $t$ consists of the goal instruction $G$, the rule specification $R$, the updated maze image $I_t$, and the accumulated reasoning trace $\mathcal{H}_{t}$.

The model generates an executable program $c_t$ that specifies which tool to invoke and how to parameterize it: \\$(r_t, c_t) = f_\theta(G, R, I_t, \mathcal{H}_{t})$
Importantly, the tools themselves are not synthesized during inference; they are pre-defined by an external LLM and made available to the controller during the spatial planning process.
Executing $c_t$ invokes a tool $\tau_t \in \mathcal{T}$ and returns an observation $o_t$, which is incorporated into $\mathcal{H}_{t+1}$ for the next reasoning step.
For example, as illustrated in the lower part of Fig.~\ref{fig:method}, when invoking a rule verification tool $\tau_{\text{ver}}$, the executable code $c_t$ specifies a call that takes the current partial action sequence $\mathbf{a}_{0:t}$ as input. The tool evaluates whether $\mathbf{a}_{0:t}$ satisfies the rule constraints defined by $R$ and returns a boolean signal $o_t \in \{\texttt{true}, \texttt{false}\}$. This feedback allows the model to explicitly verify rule compliance and adjust subsequent action decisions accordingly.

This process continues until a termination condition is met, producing an action sequence $a = (a_1, a_2, \dots, a_T),$ such that executing $a$ from the start state reaches the target and satisfies all rule constraints. By explicitly modeling planning as a sequence of reasoning and tool-invocation steps, the framework teaches the model \textit{how to solve} rule-constrained spatial planning problems, rather than memorizing rule-specific behaviors.


\section{Experiment}

\subsection{Experimental Setups}

\paragraph{Training Objective.}
For each training example containing $N$ controller interaction rounds (steps of reasoning and tool invocation), we train our controller using the standard cross-entropy loss. Specifically, given the target output sequence $\mathbf{y} = \{y_1, y_2, \dots, y_L\}$ which concatenates the reasoning traces $r_t$ and executable programs $c_t$ across all steps, the training objective is to minimize:
{
\setlength{\abovedisplayskip}{6pt}
\setlength{\belowdisplayskip}{6pt}
\begin{equation}
    \mathcal{L}(\theta)
    = -\sum_{i=1}^{L}
    \log P_{\theta}(y_i \mid y_{<i}, \mathcal{H}_i, G, R, I)
\end{equation}
}
where $\theta$ denotes the model parameters. Training uses pure teacher-forced supervised fine-tuning over the complete reasoning and tool-call traces.

\paragraph{Training Setting.}
We evaluate our framework on two spatial reasoning environments: (1) \textbf{RegularMaze}, which involves conventional pathfinding under basic geometric constraints, and (2) \textbf{QuestMaze}, which extends the task by incorporating semantic objects (e.g., keys, hearts). Each environment contains 3,600 training and 400 test samples. Specifically, 100 test instances are associated with rules observed during training (seen rules), while the remaining 300 instances correspond to previously unseen rules. To further analyze performance under varying reasoning complexity, the unseen rules are evenly divided into three difficulty levels—\emph{easy}, \emph{medium}, and \emph{hard}—with 100 instances in each category. We fine-tune Qwen2.5-VL (3B)~\cite{bai2025qwen2-5vl} for 15 epochs with a batch size of 32.

\renewcommand{\arraystretch}{0.7}
\begin{table*}[h]
\centering
\caption{Model performance on RegularMaze. "Seen Rule" denotes the performance on rules encountered during training, while "Unseen Rule" covers zero-shot rule generalization across three difficulty levels. $^\dagger$ denotes the post-trained model.}
\label{tab:regularmaze}
\begin{tabular}{l|cc|cccccccc}
\toprule
\multirow{3}{*}{\textbf{Model}} & \multicolumn{2}{c|}{\textbf{Seen Rule}} & \multicolumn{8}{c}{\textbf{Unseen Rule}} \\
\cmidrule(lr){2-3} \cmidrule(lr){4-11}
& \multicolumn{2}{c|}{\textbf{Avg.}} & \multicolumn{2}{c|}{\textbf{Easy}} & \multicolumn{2}{c|}{\textbf{Medium}} & \multicolumn{2}{c|}{\textbf{Hard}} & \multicolumn{2}{c}{\textbf{Avg.}} \\
\cmidrule(lr){2-3} \cmidrule(lr){4-5} \cmidrule(lr){6-7} \cmidrule(lr){8-9} \cmidrule(lr){10-11}
& EM (\%) & PR (\%) & EM (\%) & PR (\%) & EM (\%) & PR (\%) & EM (\%) & PR (\%) & EM (\%) & PR (\%) \\
\midrule
\multicolumn{11}{l}{\textit{Proprietary Model}} \\
\midrule
Gemini 2.5 Pro & & & & & & & & & & \\
\quad - Direct & 48.0 & 52.5 & 59.0 & 63.9 & 39.0 & 45.9 & 39.0 & 45.5 & 45.6 & 51.7 \\
\quad - CoT & 47.0 & 52.2 & 64.0 & 67.3 & 46.0 & 51.0 & 44.0 & 49.3 & 51.3 & 55.8 \\
Gemini 2.5 Flash (think) & 37.0 & 41.3 & 46.0 & 51.3 & 46.0 & 49.6 & 30.0 & 35.5 & 40.6 & 45.4 \\
\midrule
\multicolumn{11}{l}{\textit{Open-Source Model}} \\
\midrule
Qwen 2.5-VL-Instruct-3B & & & & & & & & & & \\
\quad - Direct & 0.0 & 6.6 & 0.0 & 10.7 & 0.0 & 10.7 & 0.0 & 13.9 & 0.0 & 11.7 \\
\quad - CoT & 0.0 & 3.6 & 1.0 & 9.9 & 0.0 & 5.9 & 0.0 & 4.1 & 0.3 & 6.6 \\
\quad - SFT$^\dagger$ & 94.0 & 95.3 & 78.0 & 82.1 & 68.0 & 72.2 & 65.0 & 67.6 & 70.3 & 73.9 \\
DMP-3B$^\dagger$~(Ours) & \textbf{98.0} & \textbf{98.4} & \textbf{95.0} & \textbf{96.7} & \textbf{88.0} & \textbf{91.9} & \textbf{87.0} & \textbf{89.6} & \textbf{90.0} & \textbf{92.7} \\
\bottomrule
\end{tabular}
\end{table*}



\paragraph{Evaluation Metrics.}
We adopt two complementary evaluation metrics~\cite{xu2025visualplanningletsthink,yang2025vrbench} for the selected task: Exact Match (EM), and Precision Rate (PR). 
Formally, for each test sample $i$ with ground-truth trajectory length $n_i$, we define:
\begin{align}
    \text{EM}_i &= \prod_{j=1}^{n_i} \mathbb{I}(\hat{v}_{ij} = v_{ij}), \\
    \text{PR}_i &= \frac{1}{n_i} \sum_{j=1}^{n_i} \left[ \prod_{k=1}^{j} \mathbb{I}(\hat{v}_{ik} = v_{ik}) \right],
\end{align}
where $\hat{v}_{ij}$ and $v_{ij}$ are the predicted and ground-truth validity labels at step $j$. EM measures whether the complete trajectory exactly matches the shortest valid path; PR evaluates the proportion of consecutively correct steps.
All rules are presented to the model in natural language at test time. An unseen rule is one whose logical formalization is absent from the training set.

\paragraph{Baselines.}
To facilitate comparison for language-based planning, we adopt Qwen 2.5-VL-Instruct~\cite{bai2025qwen2-5vl}, on both inference-only (Direct and CoT) and post-training settings. We further evaluate multimodal
reasoning performance of proprietary models with Gemini 2.5 Pro~\cite{comanici2025gemini} and Gemini 2.5 Flash with thinking mode.

\subsection{Main Results}

\paragraph{Results on RegularMaze}

Tab.~\ref{tab:regularmaze} presents the performance comparison on the RegularMaze benchmark, evaluating both in-distribution (seen rules) and out-of-distribution (unseen rules) generalization. We observe that: 
(1) \textit{Overall Performance:} DMP-3B achieves the best results across all metrics. It reaches 98.0\% EM and 98.4\% PR on seen rules, and 90.0\% EM and 92.7\% PR on unseen rules, substantially outperforming all baselines. Compared to Qwen2.5-VL with supervised fine-tuning, DMP improves unseen-rule EM from 70.3\% to 90.0\%, demonstrating a significant gain in zero-shot rule generalization.
(2) \textit{Seen Rule Performance:} On seen rules, SFT already achieves strong performance (94.0\% EM), while DMP further improves it to 98.0\%. This suggests that disentangled planning not only generalizes better but also reduces error accumulation even in in-distribution settings.
(3) \textit{Unseen Rule Generalization:} The advantage of DMP becomes more evident on unseen rules. While SFT obtains 70.3\% EM on average, DMP achieves 90.0\%, indicating substantially stronger generalization to novel constraints. In contrast, prompting-based approaches (Direct and CoT), even with proprietary models, struggle to handle rule-compliant planning, highlighting the intrinsic difficulty of the task.
(4) \textit{Performance Across Difficulty Levels:} Across different difficulty levels, all methods degrade as rule complexity increases. However, DMP maintains consistently strong performance, achieving 95.0\%, 88.0\%, and 87.0\% EM on easy, medium, and hard rules, respectively. The performance gap between DMP and SFT further widens on harder rules, suggesting that disentangled planning is particularly beneficial for handling complex compositional constraints.
(5) \textit{Effect of Text-Based Representations:} We further compare with textual planning variants in Tab.~\ref{tab:textual-planning}. Direct SFT achieves 65.3\% EM under unseen rules, and incorporating explicit representations such as coordinates (69.0\%) or ASCII grids (67.0\%) provides only marginal improvements. These results indicate that improving textual representations alone is insufficient for reliable spatial reasoning. In contrast, DMP leverages visual grounding together with executable tools, leading to substantially better performance.
Overall, the results demonstrate that DMP improves both rule compliance and generalization, especially in challenging unseen and high-complexity settings.

\renewcommand{\arraystretch}{0.7}
\begin{table*}[h]
\centering
\caption{Model performance on QuestMaze. "Seen Rule" reflects the model's ability on training rules, while "Unseen Rule" evaluates zero-shot generalization across three difficulty levels. $^\dagger$ denotes the post-trained model.}
\label{tab:questmaze}
\begin{tabular}{l|cc|cccccccc}
\toprule
\multirow{3}{*}{\textbf{Model}} & \multicolumn{2}{c|}{\textbf{Seen Rule}} & \multicolumn{8}{c}{\textbf{Unseen Rule}} \\
\cmidrule(lr){2-3} \cmidrule(lr){4-11}
& \multicolumn{2}{c|}{\textbf{Avg.}} & \multicolumn{2}{c|}{\textbf{Easy}} & \multicolumn{2}{c|}{\textbf{Medium}} & \multicolumn{2}{c|}{\textbf{Hard}} & \multicolumn{2}{c}{\textbf{Avg.}} \\
\cmidrule(lr){2-3} \cmidrule(lr){4-5} \cmidrule(lr){6-7} \cmidrule(lr){8-9} \cmidrule(lr){10-11}
& EM (\%) & PR (\%) & EM (\%) & PR (\%) & EM (\%) & PR (\%) & EM (\%) & PR (\%) & EM (\%) & PR (\%) \\
\midrule
\multicolumn{11}{l}{\textit{Proprietary Model}} \\
\midrule
Gemini 2.5 Pro & & & & & & & & & & \\
\quad - Direct & 71.0 & 77.4 & 81.0 & 86.0 & 85.0 & 88.9 & 71.0 & 74.9 & 79.0 & 83.2 \\
\quad - CoT & 74.0 & 81.5 & 86.0 & 88.5 & 80.0 & 83.8 & 70.0 & 76.8 & 78.6 & 83.0 \\
Gemini 2.5 Flash (think) & 56.0 & 63.1 & 79.0 & 84.5 & 48.0 & 57.4 & 40.0 & 50.2 & 55.6 & 64.0 \\
\midrule
\multicolumn{11}{l}{\textit{Open-Source Model}} \\
\midrule
Qwen 2.5-VL-Instruct-3B & & & & & & & & & & \\
\quad - Direct & 0.0 & 5.8 & 0.0 & 6.5 & 0.0 & 6.4 & 0.0 & 10.5 & 0.0 & 7.8 \\
\quad - CoT & 0.0 & 2.5 & 0.0 & 4.2 & 0.0 & 3.6 & 0.0 & 3.1 & 0.0 & 3.6 \\
\quad - SFT$^\dagger$ & 66.8 & 73.6 & 65.0 & 73.2 & 64.0 & 70.7 & 40.0 & 60.1 & 56.3 & 68.0 \\
DMP-3B$^\dagger$ & \textbf{91.4} & \textbf{94.3} & \textbf{92.0} & \textbf{94.3} & \textbf{88.0} & \textbf{90.7} & \textbf{84.0} & \textbf{90.1} & \textbf{88.0} & \textbf{91.7} \\
\bottomrule
\end{tabular}
\end{table*}


\paragraph{Results on QuestMaze}

Tab.~\ref{tab:questmaze} presents the results on QuestMaze, which introduces semantic objects and requires joint spatial and object-aware reasoning. We observe that:
(1) \textit{Overall Performance:} DMP-3B achieves the best performance among all methods, reaching 91.4\% EM on seen rules and 88.0\% EM on unseen rules. Compared to SFT, DMP improves unseen-rule EM from 56.3\% to 88.0\% (+31.7\%), showing strong generalization to novel rules.
(2) \textit{Comparison with Proprietary Models:} Unlike in RegularMaze, Gemini 2.5 Pro achieves relatively strong performance on QuestMaze (up to 79.0\% EM with direct prompting). This suggests that its strong language and reasoning capabilities are better suited to the cell-based layouts without complex wall structures. However, it still falls short of DMP, especially under harder settings.
(3) \textit{Unseen and Hard Rules:} As task complexity increases, the gap between DMP and SFT becomes more pronounced. On hard rules, SFT drops to 40.0\% EM, while DMP maintains 84.0\%, indicating its advantage in handling compositional constraints involving both spatial transitions and object interactions.
Overall, DMP demonstrates strong robustness and generalization in more complex environments that require both spatial and semantic reasoning.


\renewcommand{\arraystretch}{0.8}
\begin{table}[h]
\centering
\caption{Performance of text-based planning variants on RegularMaze under the "Unseen Rule" setting.}
\label{tab:textual-planning}
\begin{tabular}{lcc}
\toprule
Model & EM (\%) & PR (\%) \\
\midrule
Direct SFT           & 65.3 & 71.4 \\
w/ Coordinates       & 69.0 & 73.7 \\
w/ ASCII             & 67.0 & 70.6 \\
\bottomrule
\end{tabular}
\vspace{-1.5em}
\end{table}

\begin{figure*}[htbp]
    \centering
    \includegraphics[width=0.83\textwidth]{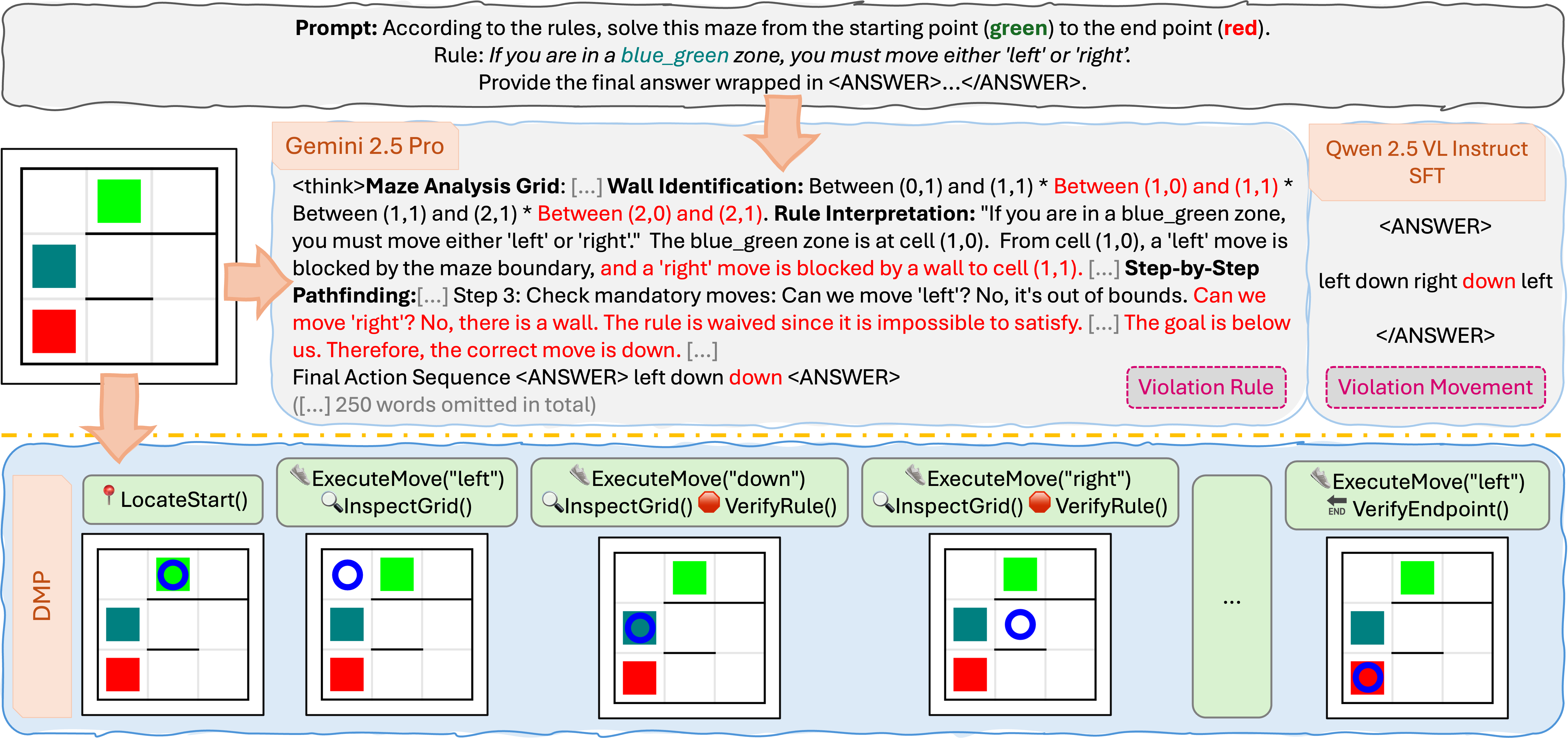}
    \caption{Visualization of a test example from RegularMaze comparing visual planning variants.}
    \label{fig:vis}
    \vspace{-0.5em}
\end{figure*}

\subsection{Qualitative Results}

Fig.~\ref{fig:vis} illustrates a representative example from RegularMaze, highlighting the differences between end-to-end textual planning and our disentangled framework under rule constraints. 
We observe distinct failure patterns in baseline methods. Qwen2.5-VL with SFT produces a seemingly reasonable action sequence, but violates the rule by executing invalid movements after entering the constrained zone. This suggests that although the model can learn general navigation behavior, it lacks an explicit mechanism to enforce rule constraints during multi-step planning, leading to unnoticed violations.
Gemini 2.5 Pro exhibits a different failure mode. It generates detailed step-by-step reasoning and correctly identifies the rule and spatial constraints. However, when encountering a situation where both allowed actions (“left” and “right”) are blocked, it heuristically concludes that the rule can be waived and proceeds with an alternative move (“down”). 
In contrast, DMP successfully generates a valid trajectory that satisfies both the goal and the rule. As shown in the bottom row of Fig.~\ref{fig:vis}, the model alternates between action execution (e.g., \texttt{ExecuteMove}) and explicit rule verification (e.g., \texttt{VerifyRule}) at each step. When a candidate action leads to a potential violation, the verification tool provides immediate feedback, enabling the model to revise its plan accordingly. This step-wise interaction ensures that all intermediate states remain rule-compliant.

\section{Conclusion}


We present RuleMaze, a benchmark for evaluating multimodal large language models (MLLMs) on rule-based visual spatial planning, a challenging task that requires the joint integration of visual perception, rule interpretation, and constrained action planning. To support scalable evaluation, we propose Language–Logic–Function Hybridization, which automatically generates and validates diverse rule sets while reducing the need for costly manual construction. Building on this benchmark, we introduce Disentangled Multimodal Planning (DMP), a structured planning framework that explicitly separates perception, plan execution, and rule verification. 
Experiments confirm that DMP significantly improves rule compliance and success over textual planning baselines. RuleMaze establishes a principled framework for advancing grounded and interpretable planning in MLLMs.

\section*{Acknowledgements}
This work was supported by the grants from the National Natural Science Foundation of China 62372014 and Beijing Nova Program.


\bibliographystyle{ACM-Reference-Format}
\bibliography{sample-base}




\end{document}


\title{Supplementary Material for Rule-Compliant Visual Spatial Planning for Multimodal Large Language Models}


\author{Yu Chen}
\affiliation{%
  \institution{Wangxuan Institute of Computer Technology, Peking University}
  \city{Beijing}
  \country{China}}
\email{yu_chen@stu.pku.edu.cn}

\author{Ting Lei}
\affiliation{%
  \institution{Wangxuan Institute of Computer Technology, Peking University}
  \city{Beijing}
  \country{China}}
\email{ting_lei@pku.edu.cn}

\author{Yaoyi Li}
\affiliation{%
  \institution{Yinwang Intelligent Technology Co., Ltd}
  \city{Shenzhen}
  \country{China}}
  \email{liyaoyi@yinwang.com}

\author{Jia Cai}
\affiliation{%
  \institution{Yinwang Intelligent Technology Co., Ltd}
  \city{Shenzhen}
  \country{China}}
  \email{caijianwpu@gmail.com} 

\author{Zhecen Wu}
\affiliation{%
  \institution{Yinwang Intelligent Technology Co., Ltd}
  \city{Shenzhen}
  \country{China}}
  \email{wuzhecen@yinwang.com}

\author{Yang Liu}
\authornote{Corresponding author.}
\affiliation{%
  \institution{Wangxuan Institute of Computer Technology, Peking University}
  \city{Beijing}
  \country{China}}
\email{yangliu@pku.edu.cn}

\renewcommand{\shortauthors}{Yu Chen et al.}

\maketitle

\appendix

\section{Limitations and Future Work}
In this work, we focus on controllable grid-based visual environments to study rule-compliant spatial planning in multimodal large language models. 
RuleMaze is intentionally designed as a controlled abstraction rather than a complete simulation of real-world environments. Since real-world settings rarely provide ground-truth trajectories under dynamically varying rules, we use synthetic data to isolate and systematically evaluate visual perception, rule interpretation, and verifiable planning.
This design allows the systematic construction of rule–maze pairs and enables precise evaluation of rule-following behaviors under varying levels of difficulty. However, such environments remain simplified abstractions of real-world spatial reasoning settings, where perceptual noise, continuous dynamics, and partially observable states are often present. While the current formulation provides a useful testbed for controlled analysis, extending the framework to more diverse and realistic visual environments remains an important direction for future work. 
At the framework level, DMP separates domain-specific visual grounding and action execution from rule-specific verification. This modularity allows adaptation to new environments to be primarily achieved by replacing the domain interface with appropriate perception and execution tools, while preserving the overall planning and verification framework.

Another limitation lies in the current formulation of rule verification. Our framework relies on externally synthesized validator functions that deterministically evaluate rule compliance. This design enables scalable dataset construction and flexible rule replacement, but assumes that rule semantics can be precisely formalized and operationalized. In broader applications, rule descriptions may be ambiguous, incomplete, or context-dependent. Developing verification mechanisms that can operate under softer or partially specified constraints may further improve the robustness of rule-based planning systems.

In addition, the proposed Disentangled Multimodal Planning (DMP) framework introduces iterative tool invocation during inference, which may incur additional computational overhead compared to direct end-to-end prediction strategies. However, it is worth noting that alternative reasoning paradigms can also exhibit substantial computational costs. For example, in our experiments, strong reasoning-based models, such as Gemini, generated over 1,500 thinking tokens in certain cases yet still failed to produce a valid rule-compliant trajectory. This observation suggests that computational cost is influenced not only by architectural design but also by the structure of the reasoning process itself. Future research may explore adaptive reasoning policies to further reduce computational overhead while maintaining reliability and interpretability.

Overall, we view RuleMaze and Disentangled Multimodal Planning (DMP) as an initial step toward systematically studying rule-compliant multimodal planning under explicit constraints. By demonstrating that models can coordinate perception, execution, and rule verification through structured interactions, this work opens new possibilities for building multimodal agents that reason under externally specified rules. Such capabilities may be particularly relevant to domains where perception and decision-making are tightly coupled, including robotics, navigation, and embodied assistance systems. We advocate for future research toward more holistic multimodal planning paradigms, where structured reasoning traces—potentially involving both symbolic and visual feedback—enable more reliable and interpretable decision-making under complex and evolving constraints.

\section{Ablation Study}

\begin{table}[t]
    \centering
    \caption{Performance of DMP-Prompting (DMP-P) on the unseen-rule split of RegularMaze. DMP-P equips each backbone with the same disentangled tools as DMP but does not involve post-training. We report the average Exact Match (EM) and Precision Rate (PR) across all unseen-rule difficulty levels. $\dagger$ denotes a post-trained model.}
    \label{tab:dmp_prompting}
    \small
    \begin{tabular}{lcc}
        \toprule
        Method & EM (\%) & PR (\%) \\
        \midrule
        Gemini-2.5-Pro             & 45.6 & 51.7 \\
        Gemini-2.5-Pro + DMP-P     & 64.0 & 67.6 \\
        GPT-5                      & 53.5 & 61.3 \\
        GPT-5 + DMP-P              & 81.4 & 84.1 \\
        Qwen2.5-VL-3B              &  0.0 & 11.7 \\
        Qwen2.5-VL-3B + DMP-P      &  0.0 & 17.3 \\
        \midrule
        DMP-3B$^{\dagger}$ (Ours)  & \textbf{90.0} & \textbf{92.7} \\
        \bottomrule
    \end{tabular}
\end{table}

\subsection{DMP-Prompting without Post-Training}
To disentangle the benefits of tool access from that of structured tool-trace training, we further introduce DMP-Prompting (DMP-P), a prompting-based variant that provides an off-the-shelf MLLM with the same perception, execution, and verification tools used in DMP, while performing no parameter updates or post-training. DMP-P follows the same iterative interaction paradigm, allowing the backbone model to inspect the visual environment, execute candidate actions, and verify rule compliance through external tools during inference. We evaluate DMP-P using Gemini-2.5-Pro, GPT-5, and Qwen2.5-VL-3B as different backbone models. Tab.~\ref{tab:dmp_prompting} reports their average EM and PR over the unseen-rule split of RegularMaze.
The results lead to three observations. First, access to the disentangled tools consistently improves rule-compliant planning, with particularly substantial gains for GPT-5, whose EM/PR increases from 53.5/61.3 to 81.4/84.1. Second, DMP-P also improves Gemini-2.5-Pro from 45.6/51.7 to 64.0/67.6, demonstrating that the proposed planning paradigm is applicable across different proprietary backbones rather than being tied to a particular model architecture. In contrast, the improvement on the smaller Qwen2.5-VL-3B backbone is limited, suggesting that tool access alone may be insufficient when the base model has not learned how to reliably select, sequence, and integrate tool outputs. Finally, DMP-3B still achieves the best EM/PR performance of 90.0/92.7, indicating that structured tool-trace training remains important for learning effective tool orchestration and generalizing to previously unseen rules.

\subsection{Tool Modules}
To better understand the contribution of each module in the proposed Disentangled Multimodal Planning (DMP) framework, we conduct a series of ablation experiments by selectively removing individual components while keeping the remaining training and evaluation settings unchanged. Results are reported in Tab.~\ref{tab:module_ablation}.

We first examine the effect of removing the execution module. Without the execution tool, the model loses the ability to explicitly update visual states after each action and relies only on text-based state updates. As a result, planning must rely on implicit reasoning over the initial observation, leading to a substantial drop in both Exact Match (EM) and Precision Rate (PR) across all difficulty levels. This observation highlights the importance of explicit state transitions for maintaining consistent spatial reasoning over multi-step trajectories.

Next, we evaluate the impact of removing the verification module. Without explicit rule checking, the model is still able to generate reasonable navigation trajectories, but lacks feedback on rule compliance. As shown in Tab.~\ref{tab:module_ablation}, performance decreases compared to the full DMP model, particularly under unseen and higher-complexity rules. This suggests that explicit verification plays an important role in preventing rule violations during long-horizon planning.

We further consider a setting where both perception and verification modules are removed. In this configuration, the model no longer has access to structured cell-state information produced by perception tools. Consequently, the verification mechanism cannot be effectively applied, as rule compliance depends on accurate intermediate state identification. As expected, this variant shows slightly lower performance than removing verification alone. This result indicates that perception-derived state representations provide the necessary context for reliable rule evaluation, and that the interaction between perception and verification modules contributes to stable planning behavior.

Overall, these ablation results suggest that the three tool modules—perception, execution, and verification—play complementary roles in the DMP framework. Among them, execution provides explicit state transitions that support consistent reasoning, verification introduces rule-aware feedback during planning, and perception enables structured interpretation of intermediate states. Removing any component leads to observable degradation, confirming the importance of disentangled tool interactions in rule-compliant spatial planning.

\begin{table*}[htbp]
\centering
\caption{Model performance on RegularMaze. "Seen Rule" denotes the performance on rules encountered during training, while "Unseen Rule" covers zero-shot rule generalization across three difficulty levels.}
\label{tab:module_ablation}
\begin{tabular}{l|cc|cccccccc}
\toprule
\multirow{3}{*}{\textbf{Model}} & \multicolumn{2}{c|}{\textbf{Seen Rule}} & \multicolumn{8}{c}{\textbf{Unseen Rule}} \\
\cmidrule(lr){2-3} \cmidrule(lr){4-11}
& \multicolumn{2}{c|}{\textbf{Avg.}} & \multicolumn{2}{c|}{\textbf{Easy}} & \multicolumn{2}{c|}{\textbf{Medium}} & \multicolumn{2}{c|}{\textbf{Hard}} & \multicolumn{2}{c}{\textbf{Avg.}} \\
\cmidrule(lr){2-3} \cmidrule(lr){4-5} \cmidrule(lr){6-7} \cmidrule(lr){8-9} \cmidrule(lr){10-11}
& EM (\%) & PR (\%) & EM (\%) & PR (\%) & EM (\%) & PR (\%) & EM (\%) & PR (\%) & EM (\%) & PR (\%) \\
\midrule
w/o Execution & 63.0 & 63.2 & 50.0 & 53.5 & 34.0 & 36.8 & 30.0 & 33.3 & 38.0 & 41.2 \\
w/o Verification & 91.7 & 92.5 & 94.0 & 94.6 & 78.0 & 79.2 & 70.0 & 73.4 & 80.6 & 82.4 \\
w/o Perception + Verification & 88.5 & 89.3 & 91.0 & 92.0 & 74.0 & 76.1 & 66.0 & 69.5 & 77.0 & 79.1 \\
DMP~(Ours) & \textbf{98.0} & \textbf{98.4} & \textbf{95.0} & \textbf{96.7} & \textbf{88.0} & \textbf{91.9} & \textbf{87.0} & \textbf{89.6} & \textbf{90.0} & \textbf{92.7} \\
\bottomrule
\end{tabular}
\end{table*}







\subsection{Training Data Scale}

We study the impact of training data scale by training both Direct SFT and DMP using 25\%, 50\%, and 100\% of the available training data. The results are summarized in Tab.~\ref{tab:data_scale_ablation}.
Empirically, we observe that performance improves consistently as the amount of training data increases for both methods. However, a clear performance gap between Direct SFT and DMP is observed across all data scales, particularly under unseen-rule settings. Notably, under the 25\% data setting, Direct SFT exhibits substantial performance degradation, whereas DMP maintains relatively stable performance. This observation suggests that the structured perception–execution–verification pipeline provides stronger inductive bias, enabling more reliable generalization even with limited supervision.
Overall, these results suggest that DMP demonstrates stronger data efficiency compared to purely text-based planning, particularly in low-data regimes and under challenging unseen-rule conditions.

\begin{table*}[htbp]
\centering
\caption{
Performance under different training data scales. We evaluate both Direct SFT and DMP using 25\%, 50\%, and 100\% of the training data. "Seen Rule" denotes performance on rules encountered during training, while "Unseen Rule" evaluates zero-shot generalization across three difficulty levels.}
\label{tab:data_scale_ablation}
\begin{tabular}{l|cc|cccccccc}
\toprule
\multirow{2}{*}{\textbf{Model}} 
& \multicolumn{2}{c|}{\textbf{Seen Rule}} & \multicolumn{8}{c}{\textbf{Unseen Rule}} \\
\cmidrule(lr){2-3} \cmidrule(lr){4-11} & \multicolumn{2}{c|}{\textbf{Avg.}}  & \multicolumn{2}{c|}{\textbf{Easy}}  & \multicolumn{2}{c|}{\textbf{Medium}}  & \multicolumn{2}{c|}{\textbf{Hard}} & \multicolumn{2}{c}{\textbf{Avg.}} \\
\cmidrule(lr){2-3} \cmidrule(lr){4-5} \cmidrule(lr){6-7} \cmidrule(lr){8-9} \cmidrule(lr){10-11}
& EM (\%) & PR (\%) & EM (\%) & PR (\%) & EM (\%) & PR (\%) & EM (\%) & PR (\%) & EM (\%) & PR (\%) \\
\midrule
Direct SFT (25\%) & 35.0 & 45.6 & 35.0 & 48.7 & 29.0 & 38.2 & 18.0 & 31.1 & 27.3 & 39.3 \\
Direct SFT (50\%) & 67.0 & 74.1 & 67.0 & 75.1 & 55.0 & 62.6 & 43.0 & 55.2 & 55.0 & 64.3 \\
Direct SFT (100\%) & 95.0 & 95.3 & 72.0 & 76.6 & 65.0 & 69.9 & 58.9 & 67.7 & 65.3 & 71.4 \\
\midrule
DMP (25\%) & 79.0 & 84.4 & 71.0 & 81.1 & 65.0 & 74.9 & 62.0 & 71.5 & 66.0 & 75.8 \\
DMP (50\%) & 95.0 & 96.5 & 90.0 & 92.9 & 86.3 & 89.4 & 84.1 & 86.5 & 86.8 & 89.6 \\
DMP (100\%) & \textbf{98.0} & \textbf{98.4} 
& \textbf{95.0} & \textbf{96.7} 
& \textbf{88.0} & \textbf{91.9} 
& \textbf{87.0} & \textbf{89.6} 
& \textbf{90.0} & \textbf{92.7} \\
\bottomrule
\end{tabular}
\end{table*}

\subsection{Text-Based Representation Variants}

To further analyze whether improved textual representations alone can enhance planning performance, we compare the proposed DMP framework with several text-based planning variants that explicitly encode spatial structure into textual outputs. The results are summarized in Tab.~\ref{tab:textual_ablation}.

We first consider a Direct SFT setting, where the model is trained to directly output the final action sequence without producing intermediate structural descriptions. This baseline achieves strong performance on seen rules, indicating that the model can learn task-specific navigation patterns under supervised training. However, performance degrades notably under unseen rules, suggesting limited generalization to novel rule compositions.

\begin{figure}
    \centering
    \includegraphics[width=0.99\linewidth]{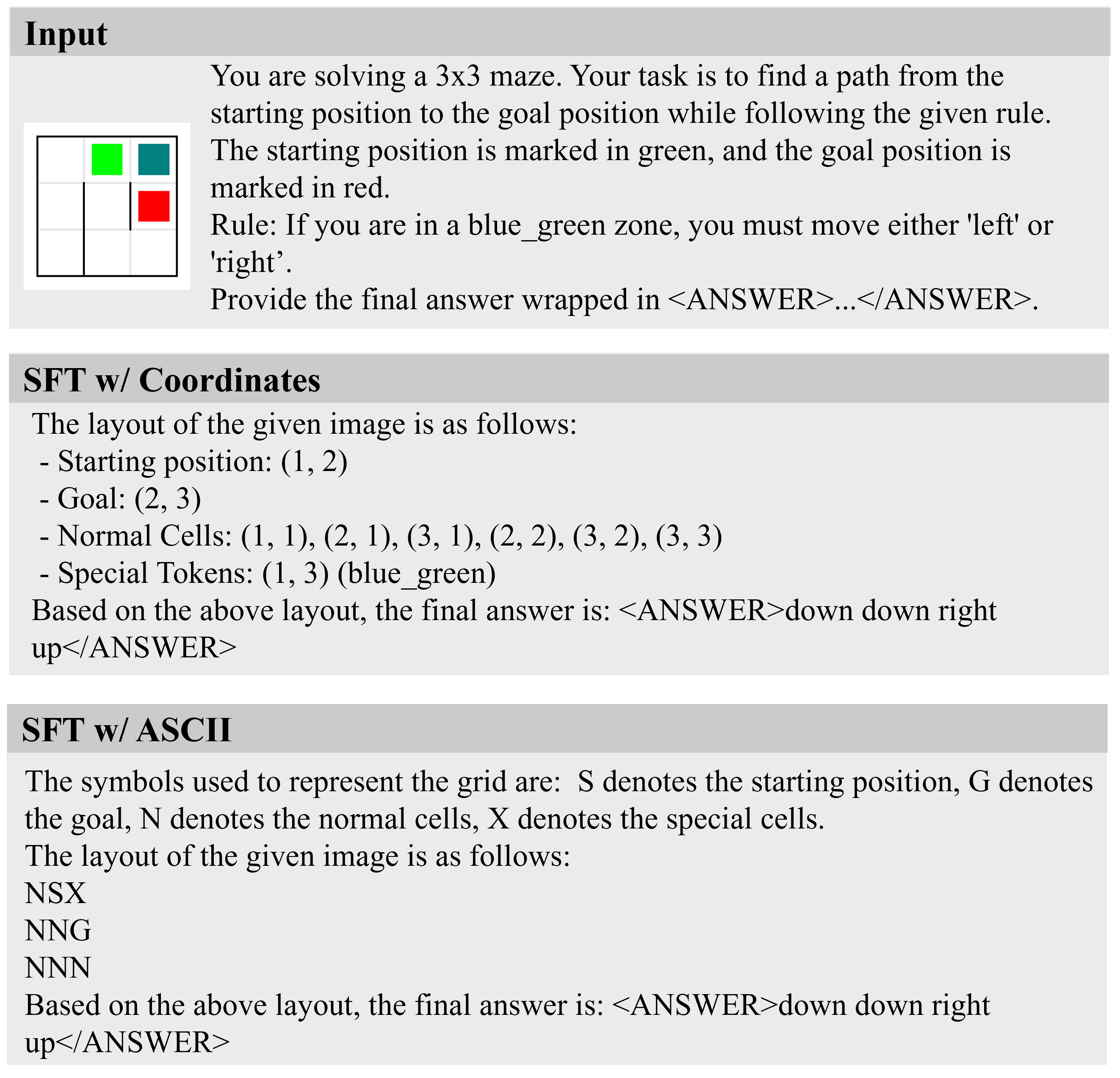}
    \caption{Examples of model outputs under different textual planning variants.}
    \label{fig:textual_planning_example}
\end{figure}

To examine whether making spatial information more explicit benefits planning, we introduce two structured textual planning
variants. In the SFT with Coordinates setting, the model is trained to first output a coordinate-based description of the grid environment, including key elements such as start and goal positions, followed by the full action sequence. In the SFT with ASCII setting, the model instead generates an ASCII-based layout of the maze before producing the action sequence. Examples of these formats are shown in Fig.~\ref{fig:textual_planning_example}.
As shown in Tab.~\ref{tab:textual_ablation}, both structured textual variants provide modest improvements over Direct SFT under unseen rules. In particular, the coordinate-based representation achieves slightly higher average performance, suggesting that explicitly encoding key spatial locations may help stabilize early planning decisions. However, the overall performance gap between these textual methods and the full DMP framework remains substantial, especially for harder rule settings.

Overall, these results suggest that enhancing textual representations alone provides limited benefits for rule-compliant visual spatial planning. While structured outputs such as coordinates or ASCII layouts can partially improve spatial awareness, they do not provide explicit mechanisms for enforcing rule constraints. In contrast, the DMP framework integrates visual grounding with executable perception and verification tools, enabling more reliable multi-step reasoning under unseen rule conditions.

\begin{table*}[htbp]
    \centering
    \caption{Performance comparison with text-based planning variants. "Seen Rule" denotes performance on rules encountered during training, while "Unseen Rule" covers zero-shot rule generalization across three difficulty levels.}
    \label{tab:textual_ablation}
    \begin{tabular}{l|cc|cccccccc}
    \toprule
    \multirow{2}{*}{\textbf{Model}} & \multicolumn{2}{c|}{\textbf{Seen Rule}} & \multicolumn{8}{c}{\textbf{Unseen Rule}} \\
    \cmidrule(lr){2-3} \cmidrule(lr){4-11}
    & \multicolumn{2}{c|}{\textbf{Avg.}} & \multicolumn{2}{c|}{\textbf{Easy}} & \multicolumn{2}{c|}{\textbf{Medium}} & \multicolumn{2}{c|}{\textbf{Hard}} & \multicolumn{2}{c}{\textbf{Avg.}} \\
    \cmidrule(lr){2-3} \cmidrule(lr){4-5} \cmidrule(lr){6-7} \cmidrule(lr){8-9} \cmidrule(lr){10-11}
    & EM (\%) & PR (\%) & EM (\%) & PR (\%) & EM (\%) & PR (\%) & EM (\%) & PR (\%) & EM (\%) & PR (\%) \\
    \midrule
    Direct SFT   & 95.0 & 95.3 & 72.0 & 76.6 & 65.0 & 69.9 & 58.9 & 67.7 & 65.3 &  71.4 \\
    w/ Coordinates  & 97.0 & 97.6 & 77.0 & 82.6 & 68.0 & 70.9 & 62.0  & 67.6 &  69.0 & 73.7  \\
    w/ ASCII & 95.0 & 95.0 & 73.0 & 78.3 & 68.5 & 70.7 & 59.5 & 62.8 & 67.0 & 70.6\\
    DMP~(Ours)      & \textbf{98.0} & \textbf{98.4} & \textbf{95.0} & \textbf{96.7} & \textbf{88.0} & \textbf{91.9} & \textbf{87.0} & \textbf{89.6} & \textbf{90.0} & \textbf{92.7} \\
    \bottomrule
    \end{tabular}
\end{table*}

\section{Additional Analysis of Rule Diversity}
To provide a more comprehensive characterization of RuleMaze, we analyze the diversity and structural complexity of its natural-language rules from two complementary perspectives: semantic type and logical connective usage. We additionally discuss the potential linguistic biases introduced by LLM-based rule generation.

\subsection{Semantic Diversity}
\label{sec:semantic_diversity}

We first categorize the rules according to four high-level semantic properties: direction dependence, history dependence, requirements, and prohibitions. A direction-dependent rule constrains the agent's movement direction based on the current state or local environment. A history-dependent rule requires information about previous states or actions, such as whether a region has previously been visited. Requirement rules prescribe actions or states that must occur, whereas prohibition rules specify actions or states that are not allowed.
As shown in Tab.~\ref{tab:rule_semantic_types}, RuleMaze contains substantial diversity across these semantic dimensions. Direction-dependent rules constitute $64.9\%$ of the rule set, while $31.3\%$ require reasoning over historical states. Requirement and prohibition constraints account for $61.3\%$ and $51.0\%$ of the rules, respectively.

\begin{table}[t]
    \centering
    \caption{Distribution of rules across different semantic types. The categories are not mutually exclusive, as a rule may exhibit multiple semantic properties.}
    \label{tab:rule_semantic_types}
    \small
    \resizebox{\columnwidth}{!}{
        \begin{tabular}{lcccc}
            \toprule
            Semantic type & Direction-dependent & History-dependent & Requirement & Prohibition \\
            \midrule
            Ratio & $64.9\%$ & $31.3\%$ & $61.3\%$ & $51.0\%$ \\
            \bottomrule
        \end{tabular}
    }
\end{table}

\subsection{Logical Structure}
\label{sec:logical_structure}

We next analyze the logical structure of the generated rules. Tab.~\ref{tab:logical_connectives} reports the average number of occurrences of different logical connectives and constraint markers per rule. Conditional and mandatory constructions are the most frequent: each rule contains, on average, $0.42$ occurrences of \textit{if} and $0.55$ occurrences of \textit{must}. Conjunction, disjunction, negation, and explicit prohibition are also represented, enabling rules to express interactions among multiple predicates and constraints.

\begin{table}[t]
    \centering
    \caption{Average number of logical connectives and constraint markers per rule.}
    \label{tab:logical_connectives}
    \small
    \resizebox{\columnwidth}{!}{
        \begin{tabular}{lcccccc}
            \toprule
            Connective
            & \textit{if} & \textit{must} & \textit{and} & \textit{or} & \textit{not} & \textit{cannot} \\
            \midrule
            Avg.\ count & 0.42 & 0.55 & 0.16 & 0.06 & 0.17 & 0.16 \\
            \bottomrule
        \end{tabular}
    }
\end{table}

\subsection{Potential Bias in LLM-Generated Rules}
\label{sec:rule_generation_bias}

Although the Language--Logic--Function pipeline supports the scalable construction of logically diverse rules, it may inherit biases from the LLM used during rule ideation. In particular, the model may repeatedly use similar concise wording patterns for rules that share related logical structures. As a result, the logical and semantic diversity quantified above does not necessarily imply equally broad linguistic diversity at the surface-form level. Models may therefore benefit partially from recurring lexical or syntactic patterns, even when the underlying logical formalizations are absent from training data.

This limitation does not affect the executable correctness of the benchmark after validator verification, but it may influence how broadly the resulting performance generalizes to naturally occurring rule descriptions. Future extensions could increase linguistic diversity through controlled paraphrasing, human-authored rules, adversarial rewriting, and evaluation on rules collected from real-world instructions. Such extensions would help distinguish generalization to unseen logical structures from robustness to diverse natural-language realizations of the same underlying constraint.

\section{Error Analysis}

\begin{figure}
    \centering 
        \includegraphics[width=0.5\textwidth]{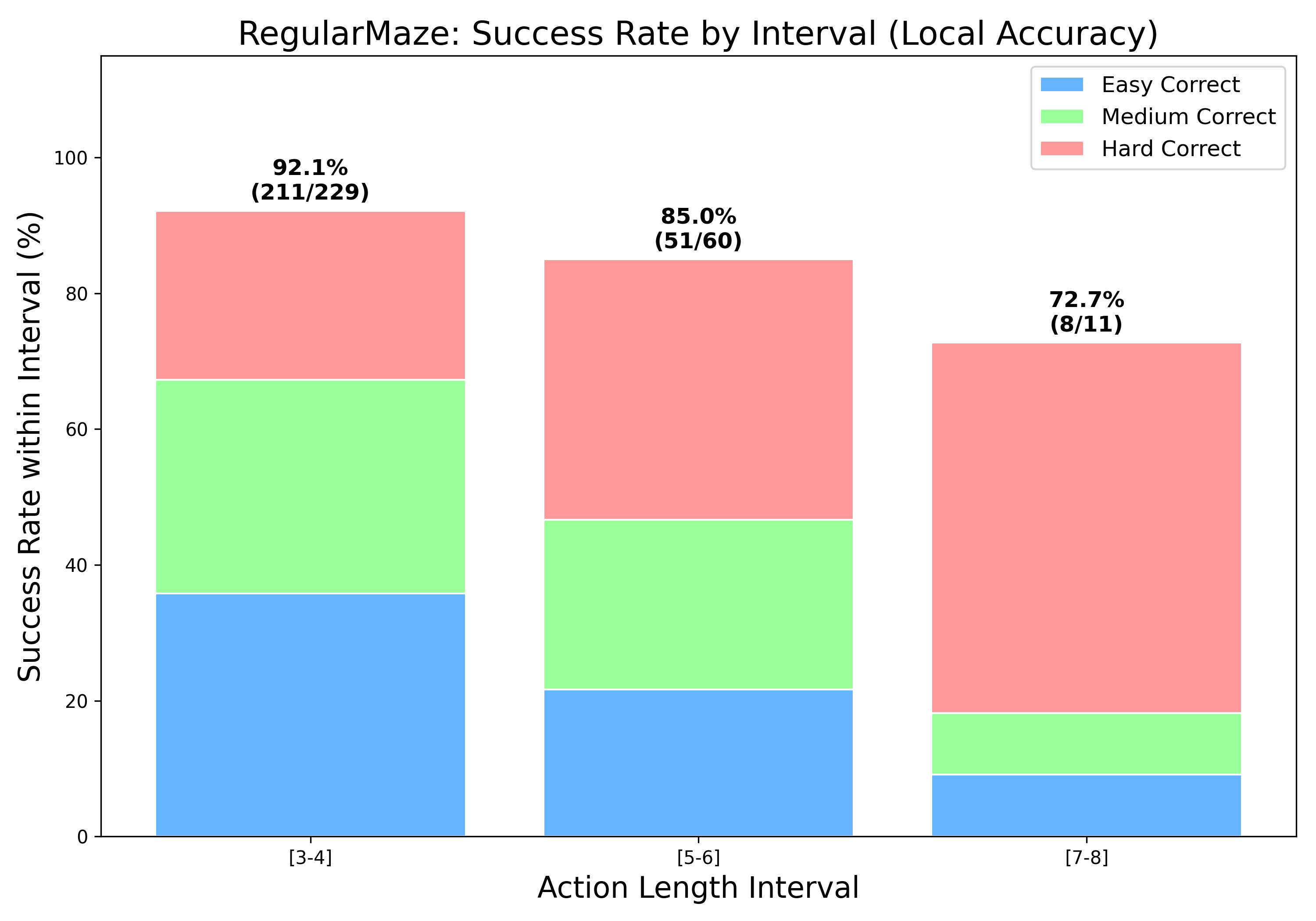} 
        \caption{Success rate by action length interval on the RegularMaze test set. Statistics are categorized by action length intervals ([3-4], [5-6], [7-8]) and further partitioned by difficulty levels (Easy, Medium, and Hard).}
        \label{fig:success_rate_by_length}
\end{figure}

To better understand the remaining failure patterns of the proposed DMP framework, we analyze prediction errors on unseen-rule settings.
Among the failure cases, approximately 37.5\% of errors are associated with missing verification calls in the proper step. In these cases, the model generates an action sequence without properly invoking the \texttt{VerifyRule()} tool, which prevents explicit detection of rule violations. This observation suggests that consistent usage of verification plays an important role in maintaining rule compliance during multi-step planning.
For the remaining cases where verification is present, we analyze the earliest step at which trajectory deviations occur. A clear trend emerges that most failures originate from early-stage decisions. Specifically, over 70\% of first errors occur at the initial step, followed by roughly 20\% at the second step and fewer than 10\% at later steps. This result indicates that trajectory correctness is highly dependent on early-step decisions, and incorrect initial actions directly determine the final mismatch outcome.
We further examine the types of rules most frequently associated with failures. These errors are commonly observed in rules that involve conditional transitions dependent on both spatial location and recent action history. Typical examples include rules that enforce directional landing constraints (e.g., requiring movement into a specific zone after a particular direction) or rules that impose short-term sequential dependencies.  
We also analyze how trajectory success varies with action sequence length. As shown in Fig.~\ref{fig:success_rate_by_length}, success rates gradually decrease as trajectory length increases, dropping from approximately 92\% for short sequences to around 73\% for longer sequences. This trend suggests that longer planning horizons introduce additional challenges for maintaining consistent rule-compliant visual spatial planning.

Overall, these observations suggest that remaining errors primarily stem from two factors: incomplete verification usage and early-stage decision deviations under complex conditional rules. These findings indicate that improving verification consistency and strengthening early-step decision reliability may further enhance the robustness of rule-compliant multimodal planning.

\section{Implementation Details}

\subsection{Training Details}
We fine-tune Qwen2.5-VL-3B-Instruct using supervised fine-tuning with LoRA adaptation. Following a parameter-efficient setup, LoRA is applied to all trainable modules with rank 8. 
The model is trained for 15 epochs with a batch size of 32. We use a learning rate of $1\times10^{-4}$ with cosine decay and a warmup ratio of 0.1. Mixed-precision training with bf16 is enabled to improve training efficiency.

\subsection{Data Generation Pipeline}

\subsubsection{Stage 1: Rule Ideation and Logical Formalization}
\label{stage:1}

\begin{figure*}[htbp]
    \centering
\includegraphics[width=0.99\textwidth]{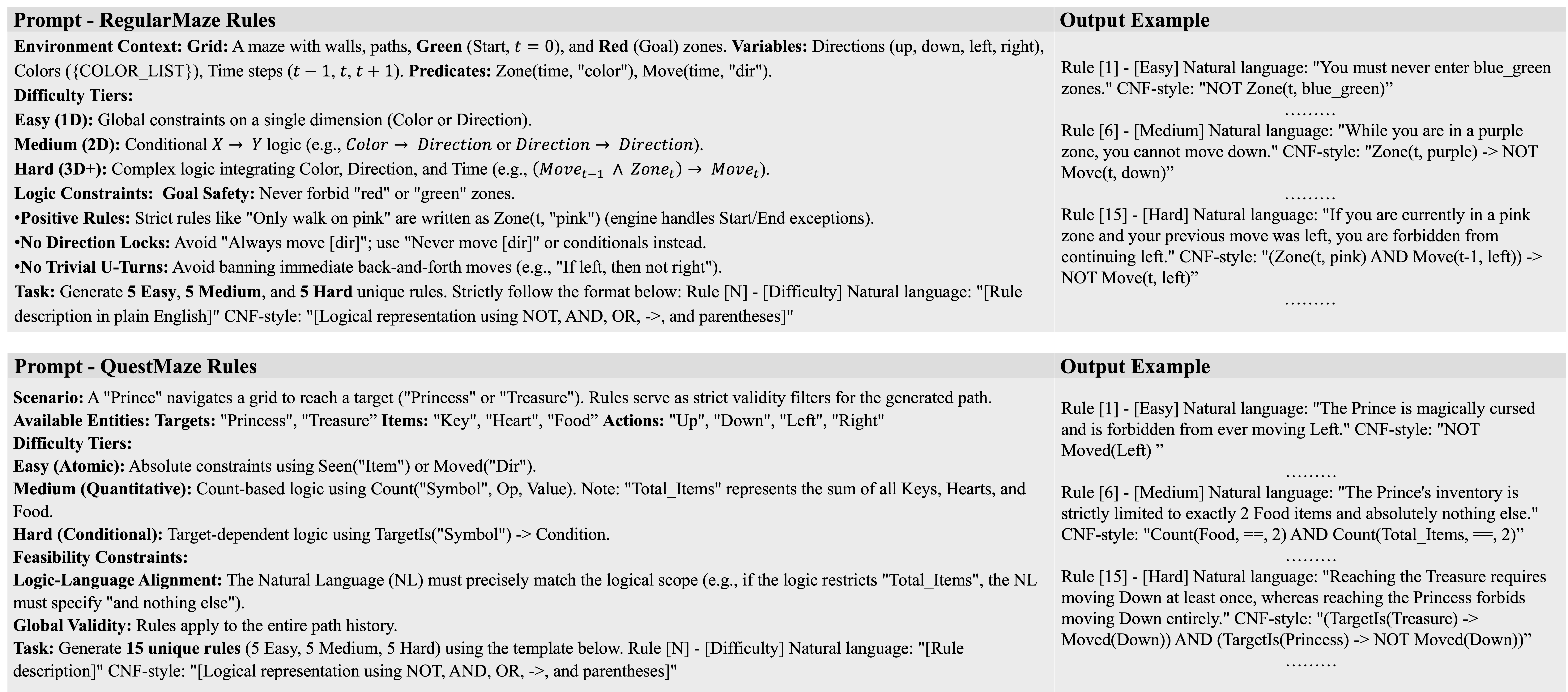}
    \caption{Example of the LLM prompt and structured output for Rule Ideation. The prompt defines logical predicates and difficulty constraints, while the output provides natural language rules paired with their symbolic CNF representations.}
    \label{fig:dataset-prompt-samples}
\end{figure*}

\paragraph{Prompting Templates}
To illustrate the logic synthesis process, Fig.~\ref{fig:dataset-prompt-samples} presents a representative prompt used for rule ideation along with its corresponding structured output. The prompt enforces strict logical constraints and formatting requirements, ensuring that the generated rules are both diverse and compatible with the setting of our task.

\paragraph{Predicate Reference}
\label{sec:predicates}
Tab.~\ref{tab:predicates} lists the full predicate vocabulary used in CNF
logical representations across both scenario types. Logical operators used across all formulas:
\texttt{NOT} (negation),
\texttt{AND} (conjunction),
\texttt{OR} (disjunction),
\texttt{->} (material implication),
and parentheses for grouping.

\begin{table}[htbp]
  \caption{CNF predicate vocabulary.}
  \label{tab:predicates}
  \footnotesize 
  \setlength{\tabcolsep}{3pt} 
  \renewcommand{\arraystretch}{1.4}
  \begin{tabularx}{\linewidth}{@{} l >{\raggedright\arraybackslash}p{2.3cm} X l @{}}
    \toprule
    \textbf{Scenario} & \textbf{Predicate} & \textbf{Semantics} & \textbf{Example} \\
    \midrule
    \textbf{RegularMaze} & \texttt{Zone(t, "col")} & Cell color at $t$ & \texttt{Zone(t, "blue")} \\
     & \texttt{Move(t, "dir")} & Action at $t$ & \texttt{Move(t-1, "up")} \\
    \midrule
    \textbf{QuestMaze} & \texttt{Seen("item")} & Item on path & \texttt{Seen("key")} \\
     & \texttt{Moved("dir")} & Direction taken & \texttt{NOT Moved("down")} \\
    & \texttt{Count(s, op, v)} & Count constr. & \texttt{Count("food", $\ge$, 3)} \\
    & \texttt{TargetIs(tg)} & Goal reached & \texttt{TargetIs("treasure")} \\
    \bottomrule
  \end{tabularx}
\end{table}

\subsubsection{Stage 2: Automated Rule Compilation}
\label{subsubsection: rule}
\begin{figure*}[htbp]
    \centering
\includegraphics[width=0.99\textwidth]{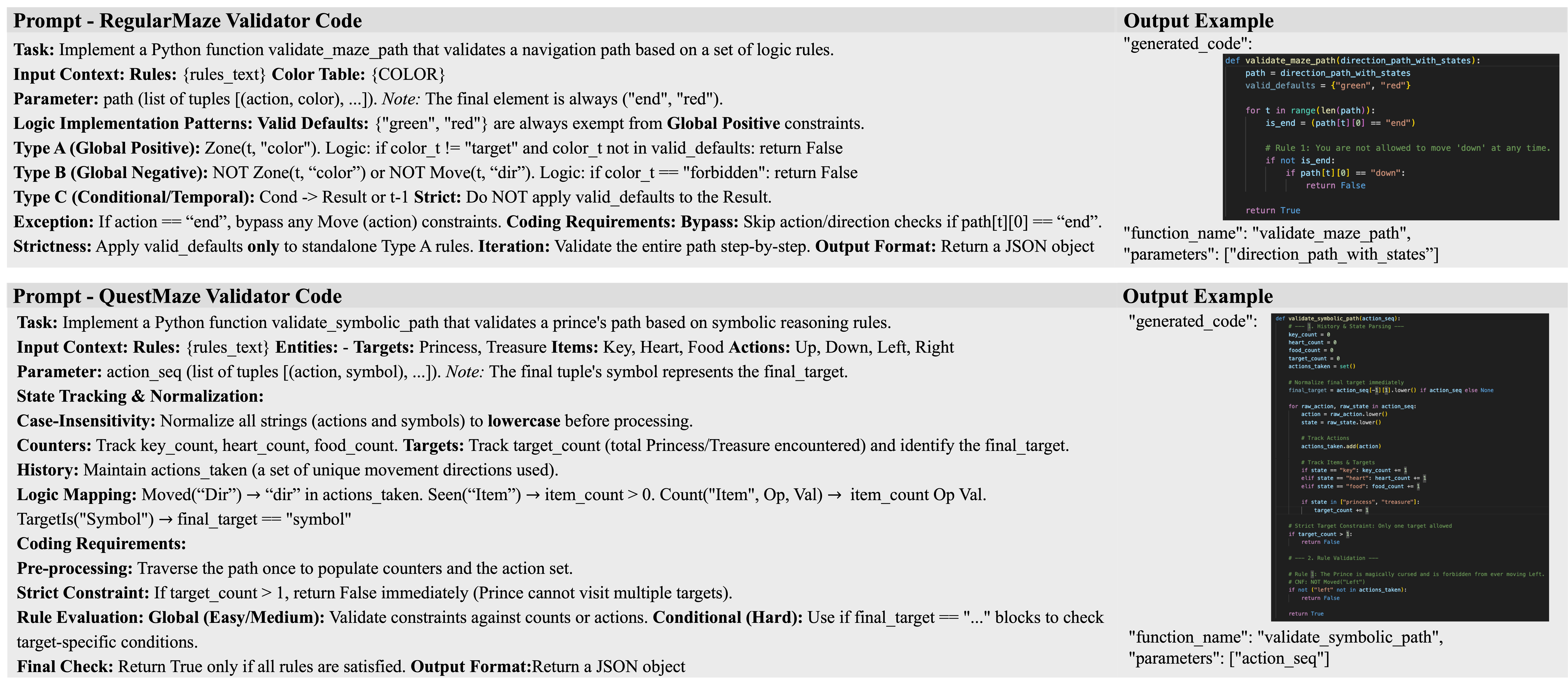}
    \caption{Example of the prompt-to-code compilation process. The system translates formalized CNF-style logic into standalone Python validator functions, ensuring deterministic rule evaluation.}
    \label{fig:code-prompt-samples}
\end{figure*}

\paragraph{Prompting Templates}
Fig.~\ref{fig:code-prompt-samples} demonstrates the automated translation from symbolic logic to executable Python code. In this stage, the system prompts the LLM to interpret the CNF-style formulas generated in Stage 1(Sec.~\ref{stage:1}) and implement them as deterministic validator functions. This process ensures that each natural language rule is backed by a verifiable programmatic ground truth.
\subsubsection{Stage 3: Maze Pool Generation}
\paragraph{Generation Detail}
Each maze is an $N{\times}N$ grid (default $N{=}3$) whose cells are connected
by passages; walls between adjacent cells are either open or closed.
A fixed $\var{loopPercent}$~$= 10$ is used: after DFS perfect-maze generation,
10\% of candidate walls are removed to introduce additional solution paths.

The hierarchical generation process is organized as follows: Alg.~\ref{alg:pool} defines the top-level loop governing the entire dataset collection. Within this framework, the procedure for synthesizing an individual maze instance is detailed in Alg.~\ref{alg:maze_instance}, which comprises two primary stages. First, Alg.~\ref{alg:create} establishes the underlying structural topology and path connectivity. Subsequently, Alg.~\ref{alg:color} specifies the heuristic for populating the maze with colored cells or symbols, ensuring a diverse distribution of semantic symbols essential for subsequent visual reasoning tasks.



\begin{algorithm}[htbp]
\caption{Maze Pool Generation}
\label{alg:pool}
\SetKwInOut{Input}{Input}\SetKwInOut{Output}{Output}
\Input{%
  Grid size $N$;
  total mazes $M$ (default $50\,000$);
  colour list $\mathcal{C}$;
  number of parallel processes $P$ (default 10)
}
\Output{Maze pool stored on disk}
\BlankLine
$\mathcal{G} \leftarrow \{(i,j) \mid 1 \le i,j \le N\}$
\tcc{all valid cell positions}

Partition $[0, M)$ into $P$ equal sub-ranges
$[c_0, c_1), [c_1, c_2), \ldots, [c_{P-1}, M)$\;
\ForEach{sub-range $[c_k,\, c_{k+1})$ \textbf{in parallel}}{
  $\func{GenerateMazeBatch}(c_k,\; c_{k+1},\; \mathcal{G},\; N,\; \mathcal{C})$\;
}
Wait for all $P$ processes to complete\;
\end{algorithm}

\begin{algorithm}[htbp]
\caption{$\func{GenerateMazeBatch}(\var{start},\, \var{end},\, \mathcal{G},\, N,\, \mathcal{C})$}
\label{alg:maze_instance}
\SetKwInOut{Input}{Input}\SetKwInOut{Output}{Output}
\Input{Index range $[\var{start}, \var{end})$; cell set $\mathcal{G}$;
       grid size $N$; colour list $\mathcal{C}$}
\Output{JSON batch file + PNG images saved to disk}
\BlankLine
$\var{count} \leftarrow \var{start}$\;
\While{$\var{count} < \var{end}$}{
  \tcc{--- Step 1: Sample start/end cells ---}
  $(\var{src},\, \var{dst}) \sim \text{Uniform-Sample}(\mathcal{G},\; k{=}2)$\;
  \BlankLine
  \tcc{--- Step 2: Generate maze via DFS + loop removal (lp = 10) ---}
  $M \leftarrow \func{CreateMaze}(N,\; \var{src},\; \var{dst})$
  \tcc{see Alg.~\ref{alg:create}}
  \BlankLine
  \tcc{--- Step 3: Enumerate all solution paths ---}
  $\mathcal{P} \leftarrow \func{GetAllSolutionPaths}(M)$\;
  \BlankLine
  \tcc{--- Step 4: Require exactly 2 solution paths ---}
  \lIf{$|\mathcal{P}| \neq 2$}{\textbf{continue}
    \tcc*{reject; retry}}
  \BlankLine
  \tcc{--- Step 5: Paint colored cells ---}
  $M \leftarrow \func{AddColouredCells}(M,\; \mathcal{P},\; \mathcal{C})$
  \tcc{Alg.~\ref{alg:color}}
  \BlankLine
  \tcc{--- Step 6: Render and persist ---}
  $\var{img} \leftarrow \func{RenderMazeImage}(M)$\;
  Append $\{M,\; \mathcal{P},\; \var{src},\; \var{dst},\; \var{img}\}$
  to batch buffer\;
  $\var{count} \leftarrow \var{count} + 1$\;
}
Save batch buffer as JSON; save $\var{img}$ files as PNG\;
\end{algorithm}

\begin{algorithm}[htbp]
\caption{$\func{CreateMaze}(N,\, \var{src},\, \var{dst})$ --- DFS with loop injection ($\var{lp}{=}10$)}
\label{alg:create}
\SetKwInOut{Input}{Input}\SetKwInOut{Output}{Output}
\Input{Grid size $N$; start cell $\var{src}$; goal cell $\var{dst}$}
\Output{Maze map $M$ with walls and cell states}
\BlankLine
Initialise all walls closed; mark $\var{src}$ as $\const{green}$, $\var{dst}$ as $\const{red}$\;
\tcc{--- Phase 1: DFS perfect-maze generation ---}
$\var{stack} \leftarrow [\var{src}]$;\quad $\var{visited} \leftarrow \{\var{src}\}$\;
\While{$\var{stack} \neq \emptyset$}{
  $u \leftarrow \var{stack.top()}$\;
  $N_u \leftarrow \{v \mid v \text{ adjacent to } u,\; v \notin \var{visited}\}$\;
  \eIf{$N_u \neq \emptyset$}{
    $v \sim \text{Uniform}(N_u)$\;
    $\func{RemoveWall}(M,\; u,\; v)$\tcc*{open passage between $u$ and $v$}
    $\var{visited} \leftarrow \var{visited} \cup \{v\}$;\quad $\var{stack.push}(v)$\;
  }{
    $\var{stack.pop}()$\;
  }
}
\BlankLine
\tcc{--- Phase 2: Add loops by removing extra walls (lp = 10) ---}
$\mathcal{P}_0 \leftarrow \func{GetAllSolutionPaths}(M)$
\tcc*{paths through perfect maze}
$n_\text{on} \leftarrow |\text{walls on a solution path}|$;\quad
$n_\text{off} \leftarrow |\text{remaining walls}|$\;
Remove $\lfloor n_\text{on}/3 \cdot 0.1 \rfloor$ random on-path walls\;
Remove $\lfloor n_\text{off}/3 \cdot 0.1 \rfloor$ random off-path walls
\tcc*{isCyclic check avoids trivial squares}
\Return $M$\;
\end{algorithm}

\begin{algorithm}[htbp]
\caption{$\func{AddColouredCells}(M,\, \mathcal{P},\, \mathcal{C})$ --- colour injection}
\label{alg:color}
\SetKwInOut{Input}{Input}\SetKwInOut{Output}{Output}
\Input{Maze grid $M$; set of solution paths $\mathcal{P}$; colour palette $\mathcal{C}$}
\Output{$M$ with some non-default cells painted a rule-relevant colour}
\BlankLine
\lIf{$\text{Uniform}(0,1) < 0.2$}{\Return $M$
  \tcc*{20\% chance: skip colouring}}
$c^* \sim \text{Uniform}(\mathcal{C})$
\tcc*{sample a colour}
$\var{type} \sim \text{Uniform}(\{1, 2\})$
\tcc*{type-1: cross-cells; type-2: solution cells}
\BlankLine
\If{$\var{type} = 1$}{
  $\mathcal{X} \leftarrow \func{FindCrossCells}(M,\; \mathcal{P})$
  \tcc*{cells shared by $\ge 2$ distinct paths}
  $k \sim \text{Uniform}\!\left(\left\{1,\ldots,|\mathcal{X}|\right\}\right)$\;
  $\mathcal{S} \sim \text{Uniform-Sample}(\mathcal{X},\; k)$\;
}
\Else{
  $\mathcal{X} \leftarrow \func{FindSolutionPathCells}(M,\; \mathcal{P})$
  \tcc*{all cells on any solution path}
  $k \sim \text{Uniform}\!\left(\left\{1,\ldots,\lfloor|\mathcal{X}|/3\rfloor\right\}\right)$\;
  $\mathcal{S} \sim \text{Uniform-Sample}(\mathcal{X},\; k)$\;
}
\ForEach{cell $u \in \mathcal{S}$}{
  $M[u].\textit{state} \leftarrow c^*.\textit{name}$\;
  $M[u].\textit{color} \leftarrow c^*.\textit{rgb}$\;
}
\Return $M$\;
\end{algorithm}


\subsection{Validator Code Implementation Analysis}

\begin{figure}
    \centering
\includegraphics[width=0.4\textwidth]{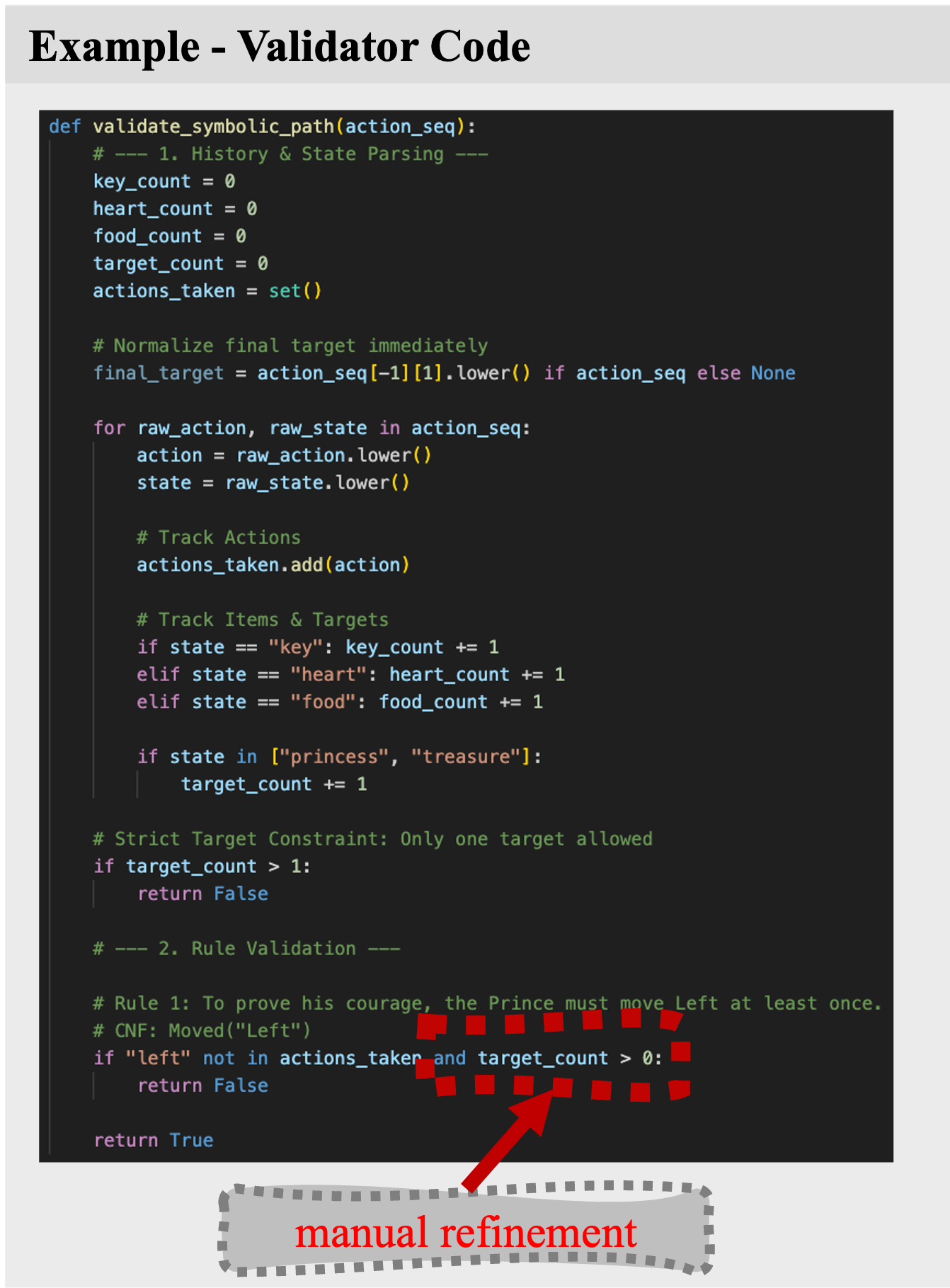}
    \caption{Implementation of the validator code with manual refinement. The red box highlights the manual refinement where the target\_count > 0 condition was added.}
    \label{fig:validator_code}
\end{figure}

Fig.~\ref{fig:validator_code} presents the core implementation of the \texttt{validate\_}\allowbreak\texttt{symbolic\_}\allowbreak\texttt{path} function. After manual refinement, this logic corresponds to the rule: "To prove his courage, the Prince must move Left at least once." When using LLMs to generate validator code, the generated implementations achieve an accuracy of 93\%, but the models frequently produce overly simplified checks, such as "left" not in actions\_taken. To fix this, we manually added a target\_count > 0 constraint to better handle two different stages: Data Matching and Dynamic Inference. In the matching stage, the system checks completed paths that have already reached the goal, so simple assertions work fine. However, during real-time inference, the system evaluates partial paths that are still in progress. Without checking if the goal has been reached, the validator would wrongly reject a path just because the agent hasn't moved "left" yet, even though it might do so in future steps. By adding the target\_count > 0 condition, we ensure that rule validation is conducted only when the agent reaches the goal state, as compliance with such rules can only be evaluated over the complete trajectory. This prevents intermediate states from being incorrectly judged as violations during the search process.

\subsection{Disentangled Tools Implementation}
We describe the five disentangled tools used in the DMP framework.
Each tool encapsulates a single atomic interaction with the maze environment,
keeping perception, execution, and verification concerns strictly separated.

The maze environment is represented as a pixel image $I_t \in \mathbb{R}^{H\times W\times 3}$.
Its internal state is \emph{not} directly accessible to the agent;
all information must be obtained by calling tools.
Tools interact with the environment through two shared primitives:
\textsc{GridDivision} parses the image to recover cell geometry
$(g_h, g_w, x_0, y_0, l_w)$, and
\textsc{CLIP} provides visual embeddings for cell-content classification.
The agent's current position is communicated across tool calls via
a colored circle marker drawn directly onto the image.

\subsubsection{Perception Tools}

\paragraph{\func{LocateStart.}}
Alg.~\ref{alg:locatestart} illustrates the workflow of the spatial localization tool, which bootstraps the agent's spatial awareness directly from raw maze images. 
The algorithm systematically scans each cell in the maze grid to extract localized patches, subsequently computing the CLIP-based cosine similarity between these patches and a reference "start" legend. 
The cell exhibiting the peak similarity score is designated as the origin. 
To maintain a stateless workflow, the tool overlays a distinctive colored circle onto the identified start cell. This visual indicator serves two purposes: first, it acts as a visual prompt for the multimodal model to perceive its current location directly from the image; second, it encodes the spatial state into the maze grid, allowing subsequent tools to decode the coordinates and update the agent's progress without relying on external memory.

\begin{algorithm}[htbp]
\caption{$\func{LocateStart}(I_t)$}
\label{alg:locatestart}
\small
\SetKwInOut{Input}{Input}\SetKwInOut{Output}{Output}
\Input{Maze image $I_t$}
\Output{Annotated image $I_{t+1}$ with a position marker at the start cell}
\BlankLine
$(g_h, g_w, x_0, y_0, l_w) \leftarrow \func{GridDivision}(I_t)$
\tcp*{infer cell geometry from image}
\BlankLine
$\sigma^* \leftarrow -\infty$;\quad $(g_x^*, g_y^*) \leftarrow (0, 0)$\;
\ForEach{cell $(g_x, g_y)$ in maze grid}{
  $\var{patch} \leftarrow \func{CropCell}(I_t,\, g_x,\, g_y)$\;
  $\sigma \leftarrow \func{CosineSim}\!\bigl(
      \func{CLIP}(\var{patch}),\;
      \func{CLIP}(I_{\mathrm{start\_legend}})
  \bigr)$\;
  \If{$\sigma > \sigma^*$}{
    $\sigma^* \leftarrow \sigma$;\quad
    $(g_x^*, g_y^*) \leftarrow (g_x, g_y)$\;
  }
}
\BlankLine
$I_{t+1} \leftarrow \func{DrawMarker}(I_t,\, g_x^*,\, g_y^*)$\;
\Return $I_{t+1}$\;
\end{algorithm}

\paragraph{\func{InspectGrid}}
Alg.~\ref{alg:inspectgrid} details the procedure for extracting the semantic content of the agent's current cell. The tool first decodes the agent's grid coordinates by localizing the marker pixel cluster within the current observation $I_t$. Once the coordinates are obtained, it extracts the corresponding cell patch and performs zero-shot classification against a predefined symbol vocabulary $\mathcal{L}$ using CLIP embeddings. Depending on the environment, $\mathcal{L}$ represents cell colors in the \textit{RegularMaze} or specific entities (e.g., \texttt{key}, \texttt{food}, \texttt{princess}) in the \textit{QuestMaze} variant. The identified symbol $s_t$ is then appended to the agent's trajectory history, providing the essential state information for subsequent evaluation by \func{VerifyRule}.

\begin{algorithm}[htbp]
\caption{$\func{InspectGrid}(I_t)$}
\label{alg:inspectgrid}
\small
\SetKwInOut{Input}{Input}\SetKwInOut{Output}{Output}
\Input{Maze image $I_t$ (with position marker at current cell)}
\Output{Semantic symbol $s_t$ of the current cell}
\BlankLine
$(g_x, g_y) \leftarrow \func{LocateMarker}(I_t)$
\tcp*{recover grid coords from marker pixels}
$\var{patch} \leftarrow \func{CropCell}(I_t,\, g_x,\, g_y)$\;
\BlankLine
\ForEach{candidate symbol $\ell$ in symbol vocabulary $\mathcal{L}$}{
  $\sigma_\ell \leftarrow \func{CosineSim}\!\bigl(
      \func{CLIP}(\var{patch}),\;
      \func{CLIP}(I_\ell)
  \bigr)$\;
}
$s_t \leftarrow \arg\max_{\ell \in \mathcal{L}}\; \sigma_\ell$\;
\lIf{$s_t \in \{\text{white},\, \text{background}\}$}{$s_t \leftarrow \texttt{normal}$}
\Return $s_t$\;
\end{algorithm}

\subsubsection{Execution Tool}

\paragraph{\func{ExecuteMove}}
Alg.~\ref{alg:executemove} defines the primary mechanism for the agent's spatial transitions within the maze. The tool first decodes the current grid coordinates from the visual marker in $I_t$ and applies a unit offset corresponding to the selected action $a_t$. It then renders the marker at the new coordinate to produce the subsequent observation $I_{t+1}$. Critically, the tool maintains a minimalist design by \textit{not} enforcing physical constraints (e.g., walls or boundaries); instead, it delegates the evaluation of rule compliance and physical validity to \func{VerifyRule}. By encoding positional updates directly as image annotations rather than internal state variables, the tool preserves a consistent image-only interface across the entire framework.

\begin{algorithm}[htbp]
\caption{$\func{ExecuteMove}(I_t,\, a_t)$}
\label{alg:executemove}
\SetKwInOut{Input}{Input}\SetKwInOut{Output}{Output}
\Input{Maze image $I_t$ with current position marker;
       action $a_t \in \{\texttt{up},\texttt{down},\texttt{left},\texttt{right}\}$}
\Output{Updated image $I_{t+1}$ with marker at new cell}
\BlankLine
$(g_x, g_y) \leftarrow \func{LocateMarker}(I_t)$\;
\BlankLine
\tcp{Apply directional offset}
$\Delta \leftarrow \bigl\{
  \texttt{up}{:}(0,{-1}),\;
  \texttt{down}{:}(0,{+1}),\;
  \texttt{left}{:}({-1},0),\;
  \texttt{right}{:}({+1},0)
\bigr\}$\;
$(g_x',\, g_y') \leftarrow (g_x,\, g_y) + \Delta[a_t]$\;
\BlankLine
$I_{t+1} \leftarrow \func{DrawMarker}(I_t,\, g_x',\, g_y')$\;
\Return $I_{t+1}$\;
\end{algorithm}

\subsubsection{Verification Tools}

\begin{algorithm}[htbp]
\caption{$\func{VerifyRule}(a_{1:t},\, s_{1:t},\, R)$}
\label{alg:verifyrule}
\SetKwInOut{Input}{Input}\SetKwInOut{Output}{Output}
\Input{Action history $a_{1:t}$; cell-state history $s_{1:t}$;
       rule $R$ (natural-language string)}
\Output{Boolean: \texttt{True} iff the trajectory complies with $R$}
\BlankLine
\tcp{Look up the synthesised Python validator for rule $R$}
$(\var{code},\, \var{fname}) \leftarrow \func{LookupValidator}(R)$\;
\lIf{$\var{code} = \varnothing$}{\Return \texttt{False}
\tcp*{unknown rule}}
\BlankLine
\tcp{Execute validator in isolated namespace}
$\tau \leftarrow \bigl[(a_i,\, s_i)\bigr]_{i=1}^{t}$
\tcp*{zip action and state histories}
$\var{ns} \leftarrow \{\}$;\quad
$\texttt{exec}(\var{code},\, \var{ns})$\;
\Return $\var{ns}[\var{fname}](\tau)$\;
\end{algorithm}

\begin{algorithm}[htbp]
\caption{$\func{VerifyEndpoint}(I_t)$}
\label{alg:verifyendpoint}
\SetKwInOut{Input}{Input}\SetKwInOut{Output}{Output}
\Input{Maze image $I_t$ with current position marker}
\Output{Boolean: \texttt{True} iff the current cell is the target cell}
\BlankLine
$(g_x, g_y) \leftarrow \func{LocateMarker}(I_t)$\;
$\var{patch} \leftarrow \func{CropCell}(I_t,\, g_x,\, g_y)$\;
\BlankLine
$s \leftarrow \func{InspectGrid}(I_t)$
\tcp*{reuse perception for cell classification}
\Return $\bigl(s = s_{\mathrm{goal}}\bigr)$\;
\end{algorithm}
\paragraph{\func{VerifyRule}.}
Alg.~\ref{alg:verifyrule} serves as the symbolic reasoning layer that bridges the natural-language rule $R$ with the agent's trajectory history, operating independently of visual input. The tool retrieves a pre-compiled Python validator function, indexed by the normalized text of $R$, which is synthesized during the dataset construction phase (cf. \textit{Validator Synthesis} in Sec.~\ref{subsubsection: rule}). The consolidated trajectory $\tau = \{(a_t, s_t)\}_{t=1}^T$ is passed to this deterministic validator, which evaluates the sequence and returns a boolean compliance status. This design enforces a strict separation of concerns: by isolating rule semantics from visual perception, the system ensures that logical verification remains invariant to visual noise, relying solely on the synthesized symbolic representation.

\paragraph{\func{VerifyEndpoint}}
Alg.~\ref{alg:verifyendpoint} determines the termination of an episode by leveraging the visual classification pipeline established in \func{InspectGrid}. The tool extracts the semantic symbol from the agent's current location in $I_t$ and compares it against the episode's designated goal symbol $s_{\mathrm{goal}}$. For \textit{RegularMaze}, $s_{\mathrm{goal}}$ is typically defined as \texttt{red}, whereas in \textit{QuestMaze} variants, it corresponds to mission-specific entities such as \texttt{princess} or \texttt{treasure}. Since $s_{\mathrm{goal}}$ is invariant throughout a given episode, the tool requires no external environmental state beyond the current visual observation. This self-contained design ensures that endpoint verification remains consistent with the framework's image-centric interface.

\begin{figure*}[htbp]
    \centering 
    \includegraphics[width=0.95\textwidth]{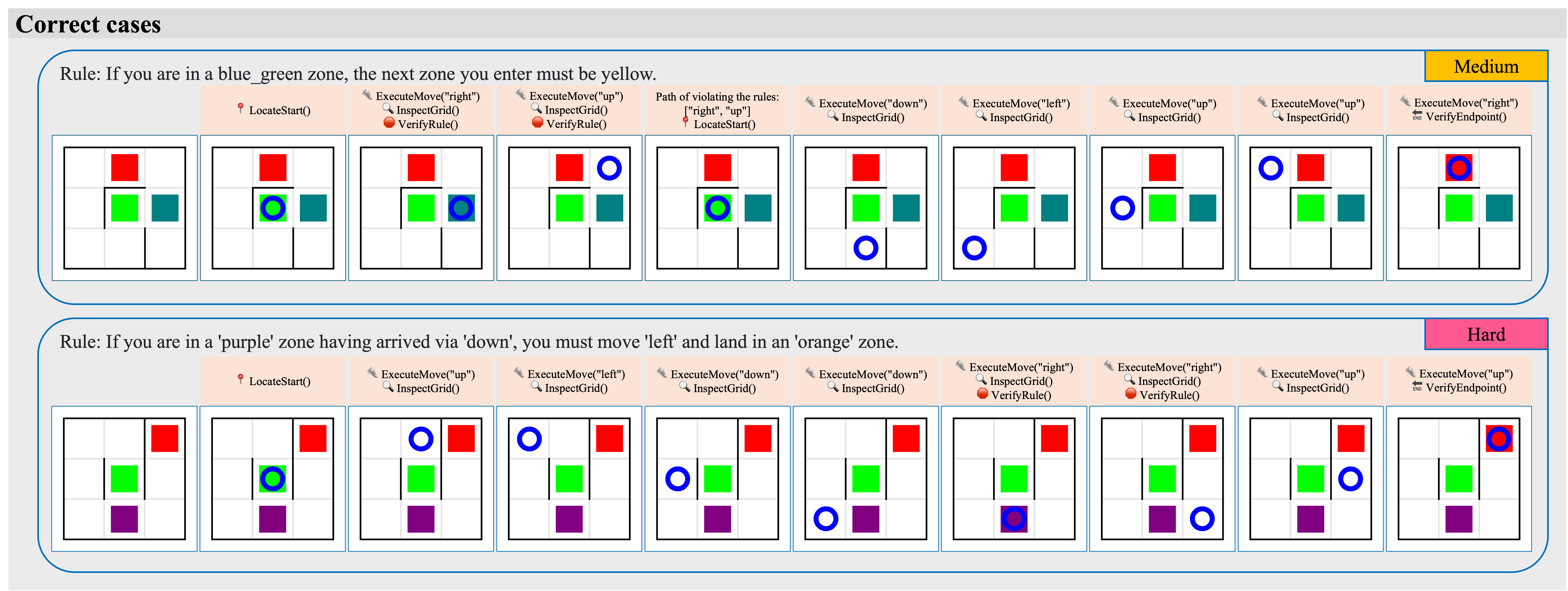}
    \caption{Examples of correct reasoning on the RegularMaze test set via DMP.}
    \label{fig:vis example regular}
\end{figure*}
\begin{figure*}[htbp]
    \centering 
    \includegraphics[width=0.95\textwidth]{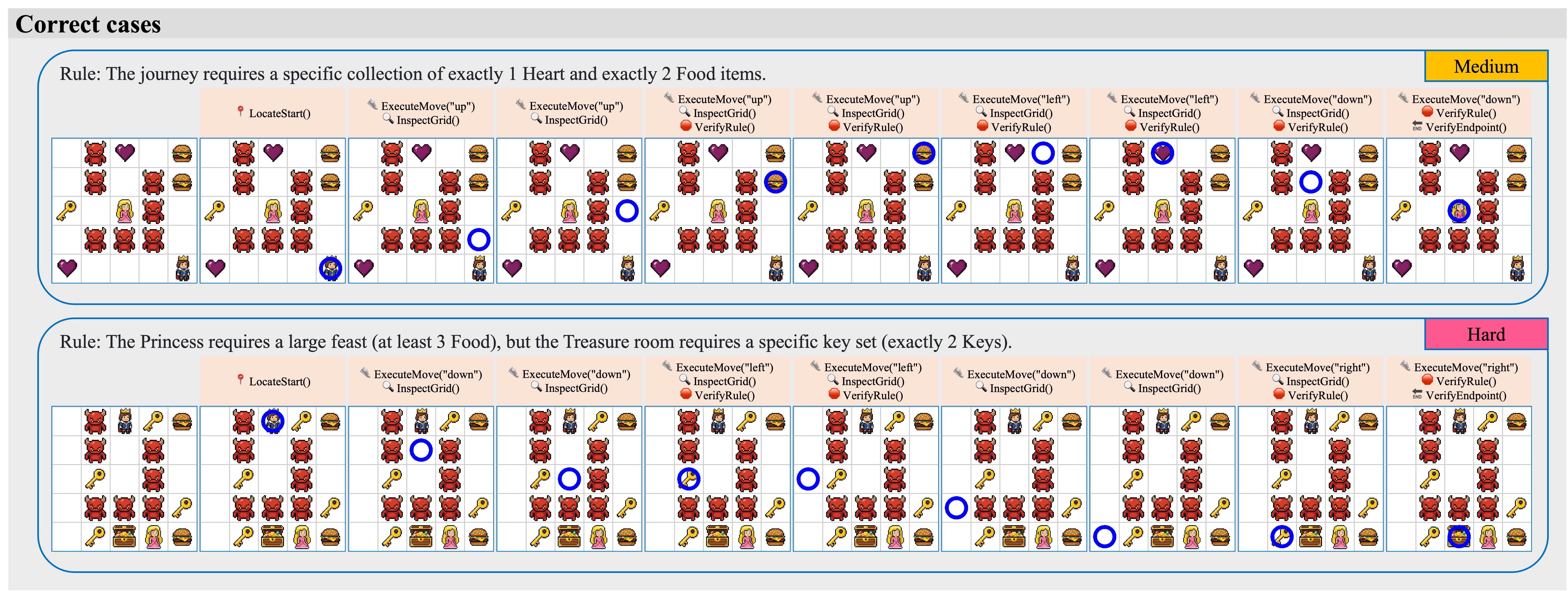}
    \caption{Examples of correct reasoning on the QuestMaze test set using DMP.}
    \label{fig:vis example quest}
\end{figure*}
\section{Qualitative Results}

We provide qualitative visualization examples to illustrate how the proposed DMP framework performs rule-aware reasoning during multi-step planning. Representative successful cases from both RegularMaze and QuestMaze test sets are shown in Fig.~\ref{fig:vis example regular} and Fig.~\ref{fig:vis example quest}, respectively.

In Fig.~\ref{fig:vis example regular}, the first-row example demonstrates how the model dynamically corrects an initially invalid decision through explicit verification. Under the rule "If you are in a blue green zone, the next zone you enter must be yellow," the model initially moves into a non-compliant zone. However, after invoking the \texttt{VerifyRule()} tool, the violation is detected, and the model subsequently revises its action choice, leading to a corrected trajectory that satisfies the rule constraint. This example highlights the role of verification feedback in enabling adaptive trajectory refinement rather than committing to early incorrect decisions.

Fig.~\ref{fig:vis example quest} further illustrates model behavior in more complex QuestMaze environments involving denser layouts and more intricate rule dependencies. Despite the increased structural complexity, the model is able to maintain consistent rule-aware reasoning across multiple steps and successfully navigate toward valid goal-reaching trajectories. These examples suggest that the proposed DMP framework remains effective even under more challenging planning conditions.

Overall, these qualitative results complement the quantitative evaluations by showing that successful trajectories are not only correct at the final step, but are also supported by interpretable intermediate reasoning behaviors, including rule checking and trajectory correction.